\documentclass[runningheads]{llncs}
\usepackage[T1]{fontenc}
\usepackage{graphicx}
\usepackage{amsmath}
\usepackage{amssymb}
\usepackage{booktabs}
\usepackage{todonotes}
\usepackage{hyperref}
\usepackage{cleveref}
\usepackage{makecell}
\usepackage{array}
\usepackage{orcidlink}

\begin{document}
\title{Multi-Objective Optimization for Synthetic-to-Real Style Transfer}
%
%
\author{Estelle Chigot \inst{1,2}\orcidlink{0009-0001-8659-6006}\thanks{Corresponding author} \and
Thomas Oberlin \inst{1}\orcidlink{0000-0002-9680-4227}\and
Manon Huguenin \inst{2}\orcidlink{0009-0007-2468-4223} \and
Dennis Wilson \inst{1}\orcidlink{0000-0003-2414-0051} }

\authorrunning{E. Chigot et al.}

\institute{Fédération ENAC ISAE-SUPAERO ONERA, Université de Toulouse, Toulouse, France \\
\email{\{estelle.chigot2, dennis.wilson, thomas.oberlin\}@isae.fr}\\
\and
Airbus, Toulouse, France\\
\email{manon.huguenin@airbus.com}}

\maketitle              

\begingroup
\renewcommand\thefootnote{}\footnotetext{%
\hspace{-2em}
\textit{Accepted in International Conference on the Applications of Evolutionary Computation (Part of EvoStar), April 2026 (EvoApplications 2026).}}
\addtocounter{footnote}{0}
\endgroup

\begin{abstract}
Semantic segmentation networks require large amounts of pixel-level annotated data, which are costly to obtain for real-world images. Computer graphics engines can generate synthetic images alongside their ground-truth annotations.
However, models trained on such images can perform poorly on real images due to the domain gap between real and synthetic images.
Style transfer methods can reduce this difference by applying a realistic style to synthetic images. Choosing effective data transformations and their sequence is difficult due to the large combinatorial search space of style transfer operators.
Using multi-objective genetic algorithms, we optimize pipelines to balance structural coherence and style similarity to target domains.
We study the use of paired-image metrics on individual image samples during evolution to enable rapid pipeline evaluation, as opposed to standard distributional metrics that require the generation of many images.
After optimization, we evaluate the resulting Pareto front using distributional metrics and segmentation performance.
We apply this approach to standard datasets in synthetic-to-real domain adaptation: from the video game GTA5 to real image datasets Cityscapes and ACDC, focusing on adverse conditions.
Results demonstrate that evolutionary algorithms can propose diverse augmentation pipelines adapted to different objectives.
The contribution of this work is the formulation of style transfer as a sequencing problem suitable for evolutionary optimization and the study of efficient metrics that enable feasible search in this space.
The source code is available at: \url{https://github.com/echigot/MOOSS}.

\end{abstract}

\thispagestyle{plain}


\section{Introduction}

Semantic segmentation assigns class labels to each pixel of an image and is an essential task for scene understanding in autonomous driving or robotics \cite{hurtado2022semantic}.
Deep learning methods achieve high performance but require large labeled datasets.
Annotating real-world images is expensive, with a single Cityscapes image requiring approximately 90 minutes of human effort~\cite{cordts2016cityscapes}.
Synthetic data from game engines like GTA5~\cite{richter2016playing} provide automatic labels, due to the engines' inherent access to object position and geometry. Yet, they induce a domain gap when models trained on synthetic images are tested on real data.

Domain adaptation techniques transfer knowledge from labeled synthetic domains to real target domains.
Recent approaches use diffusion models for style transfer~\cite{zhang2023inversion}, generating synthetic images that match real data appearance while preserving semantic structure.
However, these style transfer pipelines involve numerous steps, similar to data augmentation, and the selection and ordering of these operations remains manual.
Operators such as image darkening and sharpening, or more complex operations like image normalization through AdaIN~\cite{huang2017arbitrary}, can help transfer style from one domain to another.
As with data augmentation, these operators can be combined to improve data quality for training models on downstream tasks such as image classification.
Previous work has shown evolutionary algorithms can automate the search for data augmentation strategies in image classification~\cite{ho2019population,marc2023evolutionary,pereira2022evolving}.
In this work, we study style transfer pipeline optimization for domain adaptation, where two distributions of data must be considered.

We formulate the construction of style transfer pipelines as a multi-objective optimization task.
A pipeline consists of a sequence of transformations applied to source domain images, with each transformation altering visual properties while preserving semantic content.
We model this as a modified Traveling Salesman Problem (TSP)~\cite{flood1956traveling}, where data operators correspond to cities and a special terminal node ends the sequence.
This formulation differs from standard TSP by allowing variable-length solutions through early termination, similar to the Orienteering Problem~\cite{golden1987orienteering}.

A main challenge in optimizing style transfer pipelines is evaluation.
Metrics like FID~\cite{heusel2017fid} or CMMD~\cite{jayasumana2024cmmd} estimate distances between distributions of data but require costly generation of many images.
Training a segmentation network on each candidate dataset would be even more expensive.
Instead, we use metrics that operate on individual image pairs: DISTS~\cite{ding2022dists} for structural similarity and DreamSim~\cite{fu2023dreamsim} for perceptual style matching.
These metrics allow rapid evaluation during genetic algorithm optimization while maintaining correlation with downstream performance.

We test on domain adaptation from GTA5~\cite{richter2016playing} to Cityscapes~\cite{cordts2016cityscapes} and ACDC~\cite{sakaridis2021acdc}, which includes adverse weather conditions.
Our contributions are: formulation of augmentation pipeline selection as a combinatorial problem suitable for evolutionary optimization; identification of paired-image metrics that enable feasible optimization while correlating with distributional quality; and demonstration that multi-objective evolution produces diverse solutions representing different trade-offs between content preservation and style adaptation.

\section{Context}

\subsection{Style Transfer for Domain Adaptation}

Domain adaptation in semantic segmentation aims to train models on source domain data that generalize to target domains with different visual characteristics.
Input-level adaptation methods transform source images to resemble the target domain while preserving semantic content.
Several works have attempted to tackle the synthetic-to-real domain adaptation problem by using diffusion models to generate realistically looking datasets.

Adaptive Instance Normalization (AdaIN) provides a mechanism for style transfer by aligning feature statistics between images~\cite{huang2017arbitrary}:
\begin{equation}
\text{AdaIN}(x, y) = \sigma(y) \left(\frac{x - \mu(x)}{\sigma(x)}\right) + \mu(y)
\end{equation}
where $\mu$ and $\sigma$ denote mean and standard deviation computed across spatial dimensions for each channel.
AdaIN enables arbitrary style transfer by matching global statistics.
However, global statistics matching can produce unrealistic results.
Dominant classes in the style image may inappropriately influence other semantic classes in the content image.

Class-wise Adaptive Instance Normalization with Cross-attenTIon (CACTI) \cite{chigot2025cactif} addresses this through class-specific statistical alignment.
Rather than computing global statistics, CACTI leverages semantic segmentation masks to compute separate statistics for each semantic class.
This ensures that style characteristics transfer appropriately within semantic boundaries.

Diffusion models enable conditioning on structural information during generation.
ControlNet~\cite{zhang2023adding} adds spatial conditioning layers to pretrained diffusion models.
During the diffusion process, ControlNet processes control inputs (segmentation masks or edge maps) through an encoder that mirrors the denoising network architecture.
The control features modulate the main denoising network through zero-initialized connections, enabling structural control while preserving style generation capabilities.

DGInStyle \cite{jia2024dginstyle} first fine-tunes DreamBooth \cite{ruiz2023dreambooth} on the source domain, and then trains a ControlNet to produce images guided by segmentation maps. In the final step, they perform a style-swap procedure that replaces the source-specific UNet with a pretrained generalist UNet. This ensures that only the segmentation-based conditioning is preserved, preventing any source-style leakage during ControlNet training.
DATUM \cite{benigmim2023datum} leverages DreamBooth as well to fine-tune Stable Diffusion on a single target image for a small number of iterations. Afterwards, they prompt the model to synthesize a specific target object rendered in the style of the target image.
DoGE \cite{wang_domain_2024} computes the mean difference of CLIP embeddings pairs between the source and target dataset (Domain Gap Embedding). This embedding is then added to the latent representation of a source image, which is reconstructed using Stable UnCLIP. The approach is compatible with ControlNet, enabling dataset generation conditioned on semantic segmentation maps.

\subsection{Data Augmentation Optimization}

The optimization of data augmentation strategies has received attention as a method to improve model generalization without requiring additional labeled data. A set of operators modify data with the goal of improving model training; optimization focuses on choosing the appropriate data modification operations.

AutoAugment~\cite{cubuk2019autoaugment} introduced a reinforcement learning approach where a controller network searches over augmentation operations to find policies that maximize validation accuracy.
The method trains thousands of child models to evaluate policies, making the search computationally expensive.
RandAugment~\cite{cubuk2020randaugment} simplifies the search space by randomly selecting augmentation operations with uniform magnitude.
Fast AutoAugment~\cite{lim2019fast} employs density matching to approximate policy search, while Faster AutoAugment~\cite{hataya2020faster} uses backpropagation through differentiable augmentation implementations.
However, these methods generally require repeated model training for evaluation.

Population-based methods offer an alternative.
Population Based Augmentation (PBA)~\cite{ho2019population} addresses computational cost by learning augmentation schedules rather than fixed policies.
Pereira et al.~\cite{pereira2022evolving} apply an evolutionary strategy to discover optimal sequences of transformation functions.
Marc et al.~\cite{marc2023evolutionary} consider class-specific augmentation strategies using a genetic algorithm.
Recent differentiable methods like DADA~\cite{li2020dada} and FreeAugment~\cite{bekor2024freeaugment} enable end-to-end optimization of augmentation policies.

These policy-based formulations share a common characteristic: they determine augmentations on a per-image or per-minibatch basis during training.
This dynamic approach allows for adaptive augmentation but requires evaluating fitness through model training, which becomes prohibitively expensive when generating datasets for domain adaptation.
For style transfer in semantic segmentation, the requirements differ.
The goal is to generate a transformed dataset that can be used to train or test models that generalize to a target domain.
Rather than determining augmentations per-image during training, we seek to optimize a single pipeline that can be applied to the source domain to produce a dataset resembling the target domain.
This formulation enables the use of powerful style transfer operators such as AdaIN~\cite{huang2017arbitrary} and diffusion-based methods~\cite{zhang2023inversion}, which have greater impact on domain gap reduction than traditional geometric augmentations.
The fixed pipeline approach also enables evaluation using distributional metrics before training segmentation models.
To our knowledge, this is the first work which aims at optimizing data augmentation strategies for style transfer.

\subsection{Data Similarity Metrics}

Paired-image metrics assess similarity between individual images.
LPIPS~\cite{zhang2018unreasonable} measures perceptual similarity using features from pretrained networks.
Deep Image Structure and Texture Similarity (DISTS)~\cite{ding2022dists} measures perceptual similarity with robustness to texture variations, combining structure and texture similarity measurements:
\begin{align}
S_l(x,y) = \frac{2\mu_x\mu_y+\epsilon}{\mu_x^2+\mu_y^2+\epsilon}\\
T_l(x,y) = \frac{2\sigma_{xy}+\epsilon}{\sigma_x^2+\sigma_y^2+\epsilon}\\
\text{DISTS}(x,y) = \sum_l w_l \cdot (S_l(x,y) + T_l(x,y))
\end{align}
where $S_l$ and $T_l$ represent structure and texture similarity at layer $l$ by comparing the means $\mu_x, \mu_y$, variances $\sigma_x, \sigma_y$, and covariance $\sigma_{xy}$ of corresponding feature maps ($\epsilon$ is added to avoid instability).
Lower DISTS values indicate images share similar content.

DreamSim~\cite{fu2023dreamsim} provides a metric for evaluating perceptual similarity with focus on mid-level variations such as pose, perspective, and style.
The metric is trained on synthetic data generated using diffusion models, with human annotations of similarity judgments.
These pairwise metrics evaluate individual transformations, allowing for the evaluation of style transfer pipelines on a small number of images.

Distributional metrics like FID and CMMD provide assessment of entire generated datasets but require generating many samples.
Fr\'echet Inception Distance (FID)~\cite{heusel2017fid} computes distance between feature distributions:
\begin{equation}
\text{FID} = ||\mu_x - \mu_y||^2 + \text{Tr}(\Sigma_x + \Sigma_y - 2(\Sigma_x \Sigma_y)^{1/2})
\end{equation}
where $\mu_x$, $\mu_y$ are mean feature vectors of the two datasets and $\Sigma_x$, $\Sigma_y$ the corresponding sample covariance matrices.
CLIP Maximum Mean Discrepancy (CMMD)~\cite{jayasumana2024cmmd} addresses FID limitations by using CLIP embeddings and Maximum Mean Discrepancy with a Gaussian RBF kernel.
These distributional metrics have been used to measure the quality of style transfer methods, using distributions of generated data compared to target domain data~\cite{chung2024style}, however such a calculation would be too costly for optimization.

\section{Methodology}
\begin{figure}
    \vspace{-0.3cm}
    \centering
    \includegraphics[width=1\linewidth]{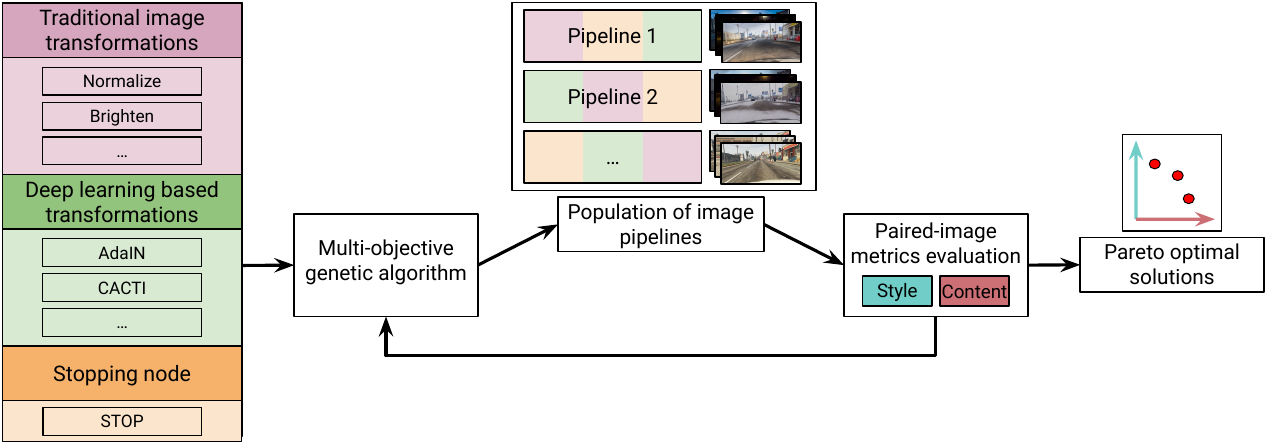}
    \caption{Overview of the proposed multi-objective pipeline optimization framework. Image transformation pipelines, composed of traditional and deep learning–based operators with a stopping node, are evolved using a multi-objective genetic algorithm. Each pipeline is evaluated using paired-image metrics that measure content preservation (DISTS) and style similarity to the target domain (DreamSim). Ultimately, the evolutionary process produces a Pareto front of non-dominated pipelines representing different trade-offs between both objectives.}
    \label{fig:diagram}
    \vspace{-0.3cm}
\end{figure}

The task of synthetic-to-real domain adaptation through style transfer can be formulated as finding a sequence of transformations $T = [t_1, t_2, \ldots, t_n]$ that, when applied to source domain images $\mathcal{S}$, produces images $\mathcal{S}'$ that maintain semantic structure while matching visual characteristics of the target domain $\mathcal{T}$.
Each transformation $t_i$ belongs to a set of available operators $\mathcal{A}$.

The problem presents two competing objectives.
First, transformed images must preserve semantic content and spatial structure to ensure ground truth annotations remain valid.
Second, transformed images should match the visual style of the target domain to reduce the domain gap.

Evaluating candidate pipelines during optimization requires balancing accuracy and computational cost.
To enable feasible optimization, we use paired-image metrics that operate on individual samples.
For each candidate pipeline, we apply the sequence to a small set of source images and compute metrics on the resulting pairs.
This reduces evaluation cost by several orders of magnitude while maintaining correlation with distributional quality and downstream performance.

We use DISTS~\cite{ding2022dists} to measure structural similarity between source and transformed images and DreamSim~\cite{fu2023dreamsim} to measure perceptual similarity between transformed images and target domain samples.
The problem becomes a multi-objective optimization task where we seek pipelines that minimize both DISTS (to preserve content) and DreamSim (to match target style).
Multi-objective optimization produces a Pareto front of non-dominated solutions representing different tradeoffs between objectives. \Cref{fig:diagram} illustrates the method.

\subsection{Traveling Salesman Problem (TSP) Formulation}

We model the augmentation pipeline selection problem as a variant of the TSP where each data operator represents a city.
The TSP seeks the shortest route visiting a set of cities exactly once and returning to the start.
We consider this suitable as the main style transfer operators used in this study, such as AdaIn~\cite{huang2017arbitrary} and ControlNet~\cite{zhang2023adding}, only operate as a single application to an image.
Other operators, such as sharpening the image or darkening it, could be applied multiple times, but do not represent the main contributors to style transfer and are thus included as one-off operator choices.

The key modification to the TSP is introduction of a terminal node $t_{\text{stop}}$ which can be visited at any point to end the pipeline.
The solution representation is a permutation of augmentation operators including the terminal node: $\pi = [\pi_1, \pi_2, \ldots, \pi_m]$ where $m = |\mathcal{A}| + 1$.
When evaluating a solution, operators are applied sequentially until the terminal node is encountered.
This is similar to the Orienteering Problem~\cite{golden1987orienteering}, which extends TSP by selecting a subset of cities to visit subject to a budget constraint, however, unlike the Orienteering Problem, we do not impose a budget on total travel time.
The discrete and combinatorial nature makes this problem appropriate for genetic algorithms.
The solution space consists of permutations of operators, a representation studied extensively in TSP contexts.
Genetic algorithms have a long and successful history of use for the TSP and variants, including the Orienteering Problem~\cite{tasgetiren2000genetic}, for a multitude of applications~\cite{potvin1996genetic,larranaga1999genetic,tsai2004evolutionary,groba2015solving,davendra2010traveling}.

\subsection{Multi-objective Evolutionary Algorithm}

We employ NSGA-II\cite{deb2002fast} to solve the multi-objective optimization problem.
NSGA-II uses non-dominated sorting to classify individuals into fronts based on dominance relationships and maintains diversity through crowding distance.

The initial population is generated using permutation random sampling, which creates valid permutations of augmentation operators and terminal node.
For crossover, we use the edge recombination crossover~\cite{whitley1991traveling}, which preserves adjacency information from parent solutions.
For mutation, we apply inversion mutation, which selects two random positions and reverses the order of all elements between them.
Duplicate elimination prevents the population from containing multiple copies of the same solution.

\section{Pipeline Optimization}

\subsection{Image Operators}

The augmentation pipeline consists of ten operations: nine transformations and one terminal node.
The transformations include six traditional image processing operations applied using Pillow, a modern version of the Python Imaging Library~\cite{pillow}.
The normalize operation adjusts color distribution to match a reference using histogram matching.
Blur applies Gaussian smoothing with a radius value of 2.0 pixels.
Brighten and darken modify image luminosity by factors of 1.5 and 0.7 respectively.
Contrast modifies contrast by a factor 1.5.
Sharpen applies unsharp masking with a factor 2.0.

Three deep learning-based style transfer operations provide more substantial domain adaptation.
AdaIN performs style transfer by aligning channel-wise statistics between content and style images.
CACTI uses class-specific statistical alignment through segmentation masks for focused style transfer.
ControlNet generates images conditioned on segmentation masks and edge maps.

For each deep learning operation, we use as reference a single set of five images, one for each weather condition considered. This ensures the method is applicable in a wide range of situations, even in the case of limited target data.

\begin{figure}
    \vspace{-0.3cm}
    \centering
    \setlength{\tabcolsep}{2pt} 
    \renewcommand{\arraystretch}{0.7} 
    \begin{tabular}{c c c}
        & Content & Style \\
        \rotatebox{90}{\parbox[c]{1.7cm}{\centering Night}} &
        \includegraphics[width=0.29\linewidth]{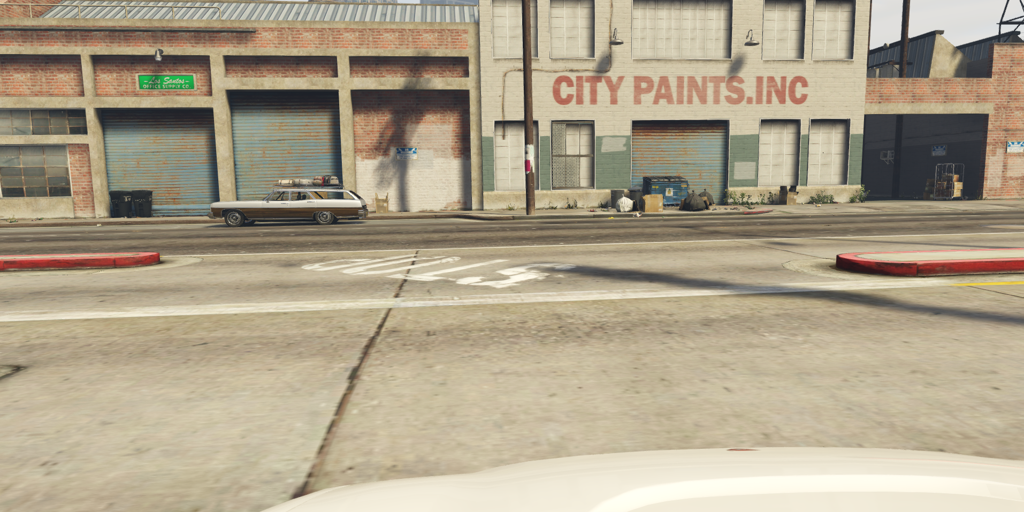} &
        \includegraphics[width=0.29\linewidth]{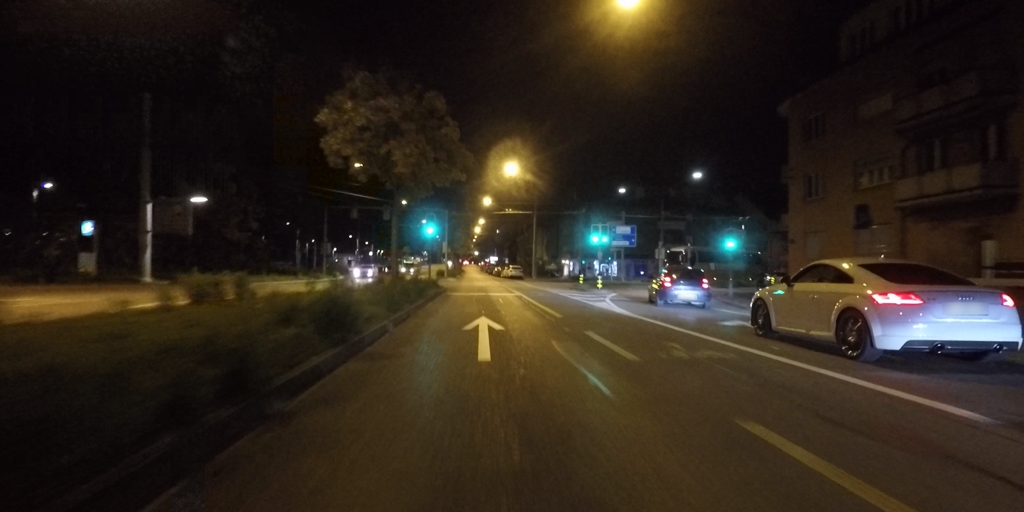} \\
        \rotatebox{90}{\parbox[c]{1.7cm}{\centering Snow}} &
        \includegraphics[width=0.29\linewidth]{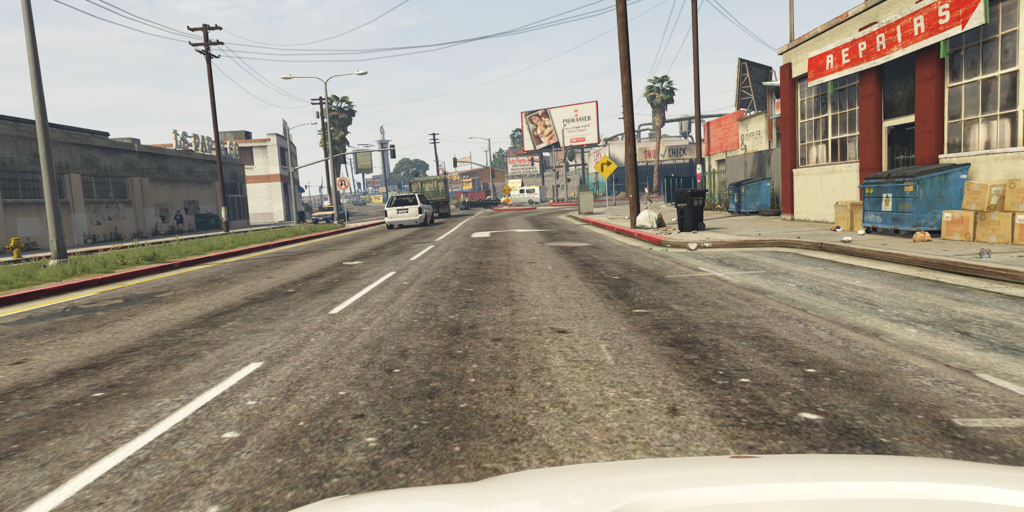} &
        \includegraphics[width=0.29\linewidth]{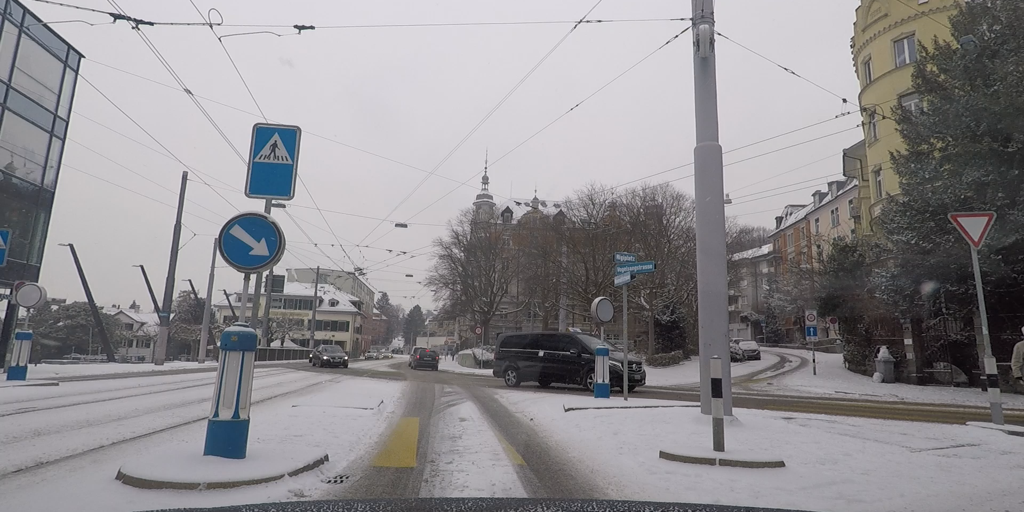} \\
        \rotatebox{90}{\parbox[c]{1.7cm}{\centering Fog}} &
        \includegraphics[width=0.29\linewidth]{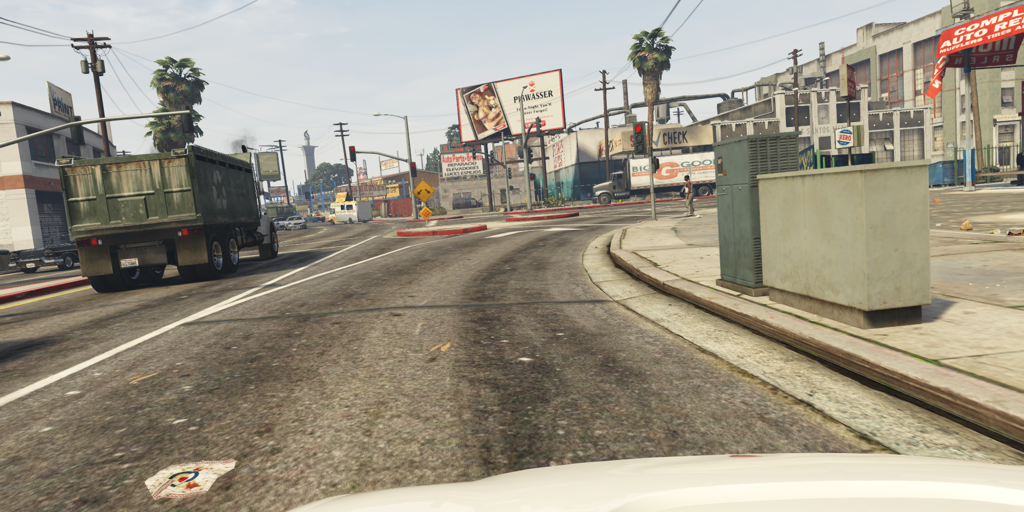} &
        \includegraphics[width=0.29\linewidth]{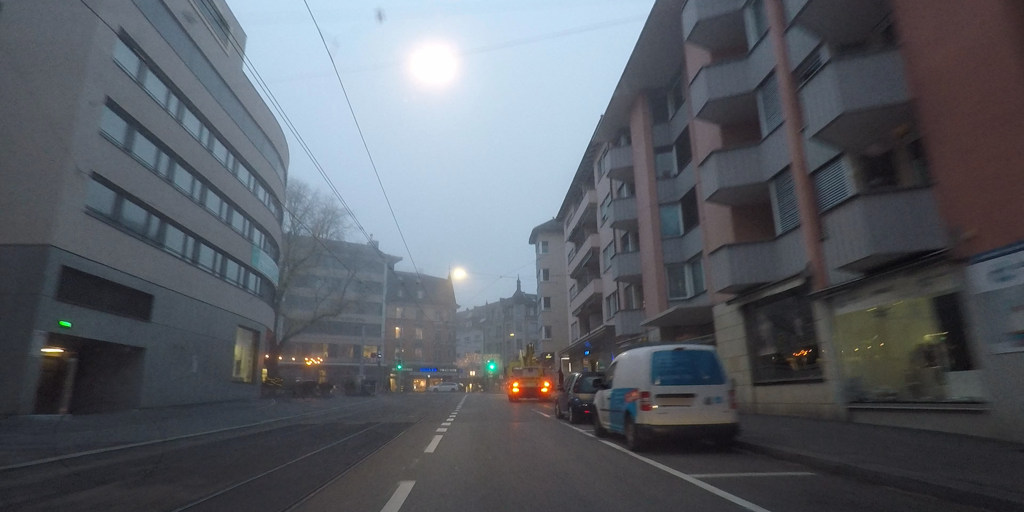} \\
        \rotatebox{90}{\parbox[c]{1.7cm}{\centering Rain}} &
        \includegraphics[width=0.29\linewidth]{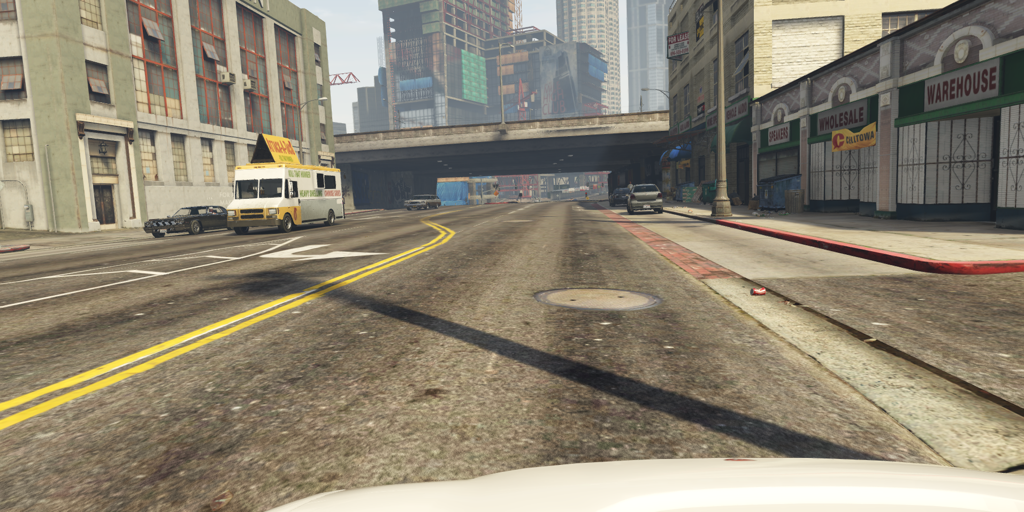} &
        \includegraphics[width=0.29\linewidth]{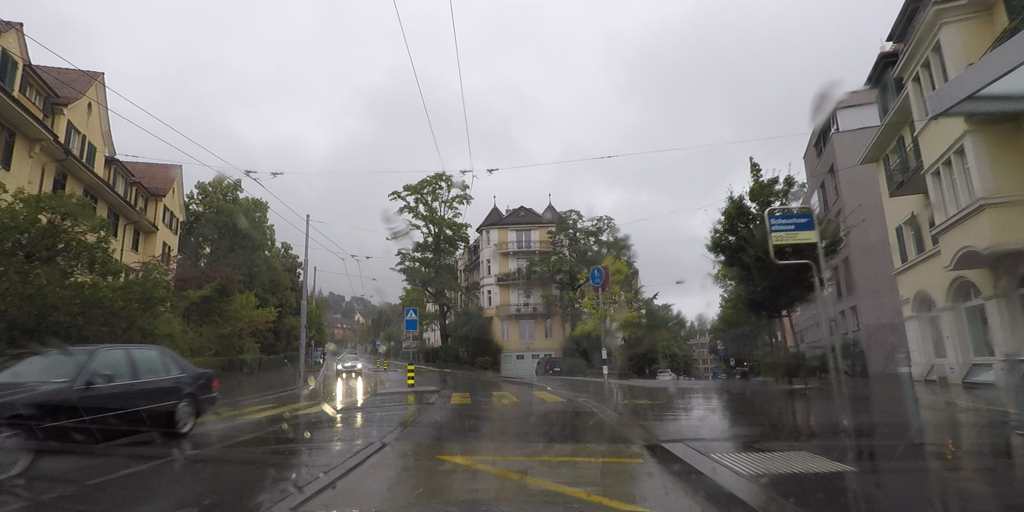} \\
        \rotatebox{90}{\parbox[c]{1.7cm}{\centering Clear Day}} &
        \includegraphics[width=0.29\linewidth]{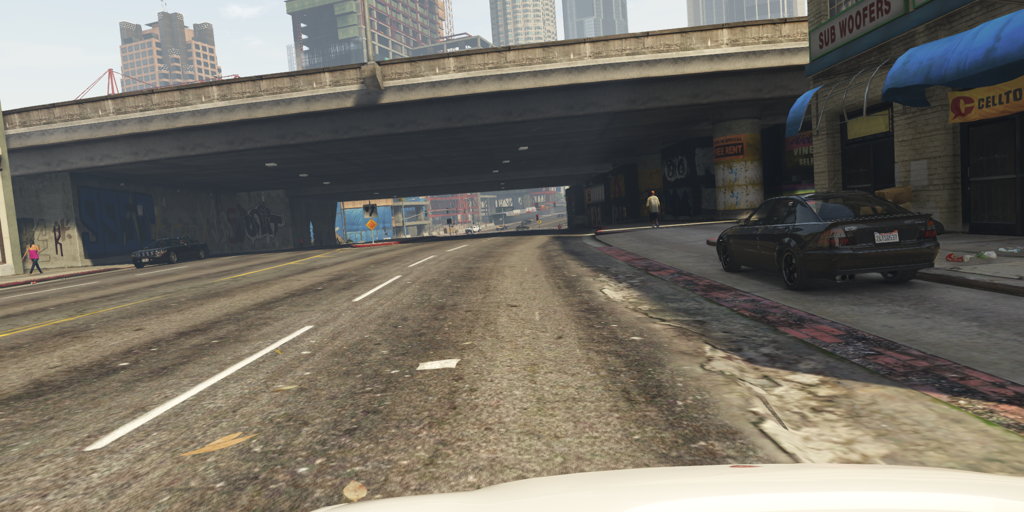} &
        \includegraphics[width=0.29\linewidth]{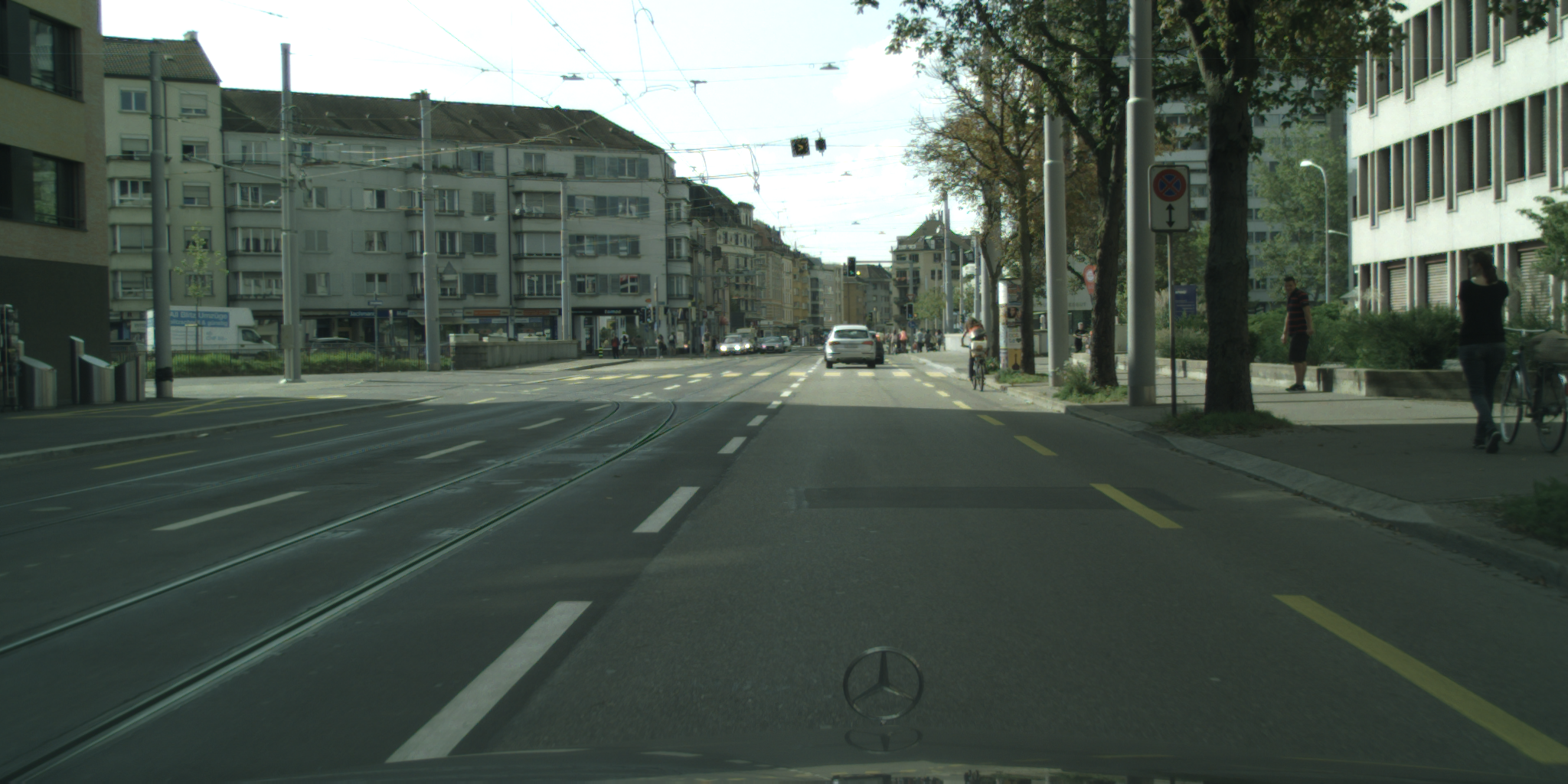} \\
    \end{tabular}
    \caption{Pairs of images (content, style) used during the optimization process, respectively used for the night, snow, fog, rain, and clear day style transfer. Content images are from the GTA5 dataset. Style images are from the ACDC dataset (night, snow, fog, and rain style) and the Cityscapes dataset (clear day style).}
    \label{fig:content_images}
\vspace{-0.3cm}
\end{figure}

\subsection{Optimization Configuration}

We use source domain images from the GTA5~\cite{richter2016playing} dataset. 
Target domain images are sampled from the Cityscapes~\cite{cordts2016cityscapes} training set for clear day conditions and ACDC~\cite{sakaridis2021acdc} training set for adverse weather conditions. Selected images from these content and style datasets are respectively shown in~\Cref{fig:content_images}.

For each candidate pipeline during optimization, we evaluate using the five sampled source images.
For each source image, we apply the pipeline transformations and measure DISTS relative to the original and DreamSim relative to selected target domain images.

Our NSGA-II configuration uses a population size of 20 individuals and generates 20 offspring per generation.
Inversion mutation and edge recombination crossover are used to generate all offspring.
The relatively small population is sufficient given the limited size of the search space.
The evolutionary process is run for 20 generations, as Pareto front convergence was observed.

Optimization was performed on a Linux server with a 32 core Intel i9 CPU and a RTX4090 GPU.
The largest time cost was image generation.
CACTI is the most expensive operator, with a maximum generation time of 40 seconds, with ControlNet in second, taking up to 2 seconds.
All other operators function in less than 1 second for generation.
The maximum evaluation time for a single pipeline was therefore roughly 200 seconds, or 3 minutes and 20 seconds on a single machine.
Full optimization took 13 hours, meaning 39 minutes per NSGA-II generation on average and 23.4 seconds per generated image.
Evaluation of chosen pipelines, requiring generation of 1250 images, also took up to 13 hours per pipeline, or 37.4 seconds per image, for solutions containing CACTI.
This demonstrates the benefit of an optimization approach using pairwise metrics over a small number of images; the full optimization took as long as the distributional evaluation of a single pipeline.
Model training on generated images using the same hardware took 5.5 hours and was performed 3 times for averages.

\subsection{Optimization Results}

\begin{figure}[t]
    \vspace{-0.3cm}
\centering
    \includegraphics[width=0.3\linewidth]{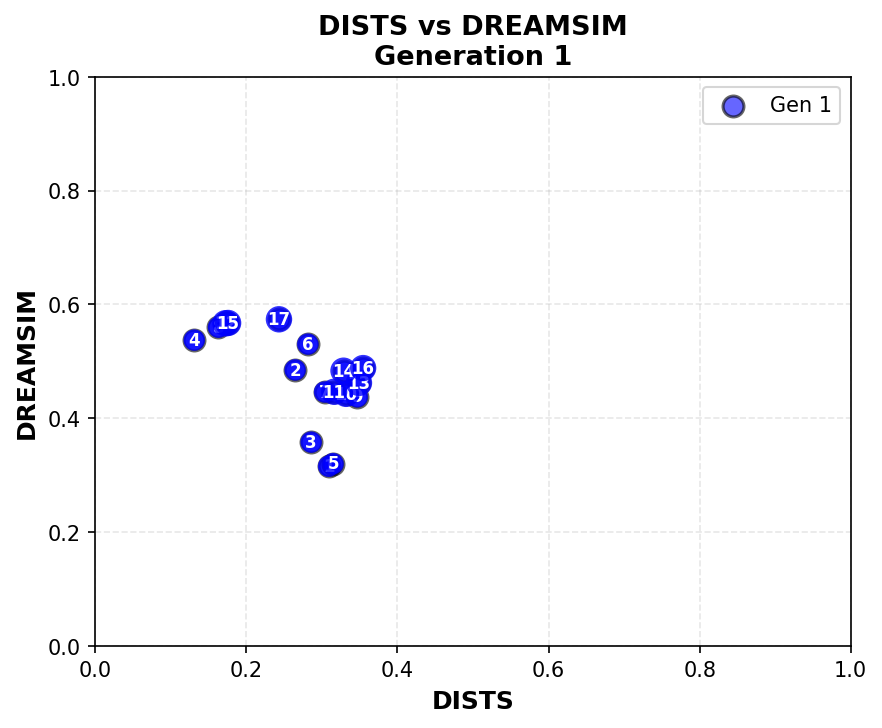}
    \includegraphics[width=0.3\linewidth]{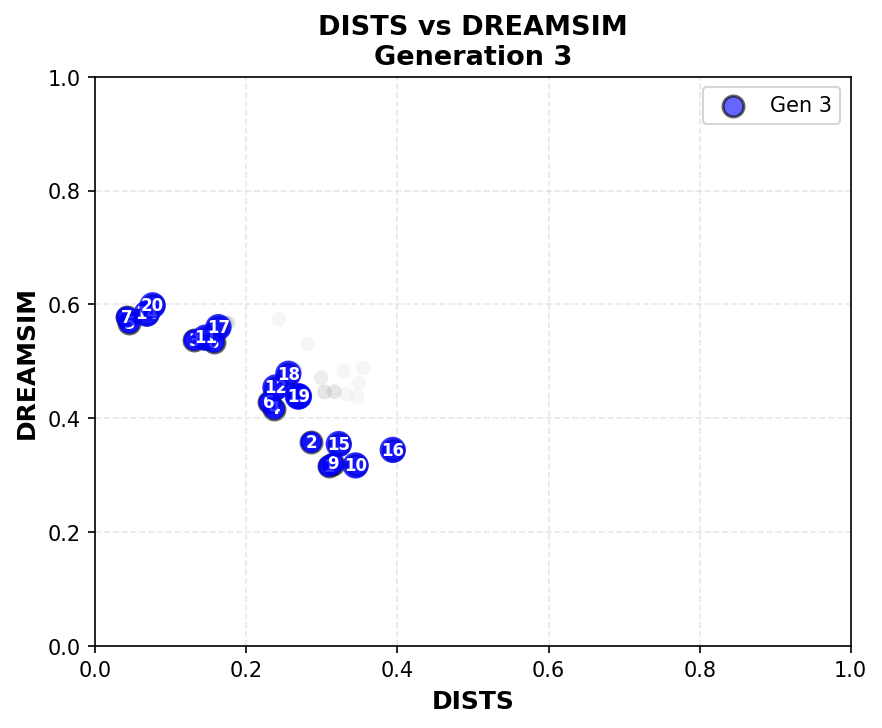}
    \includegraphics[width=0.3\linewidth]{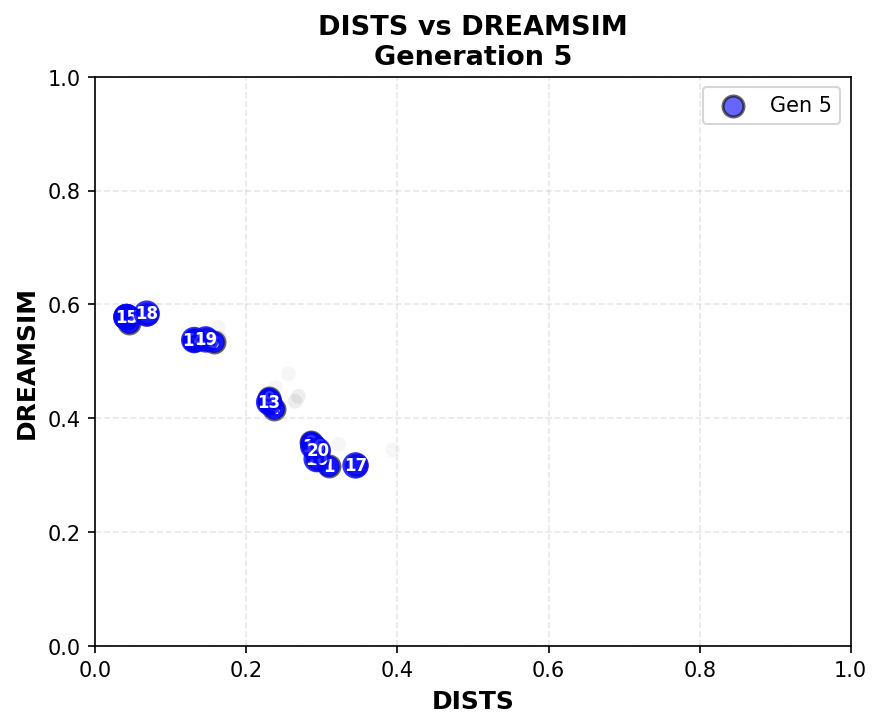}
    \includegraphics[width=0.3\linewidth]{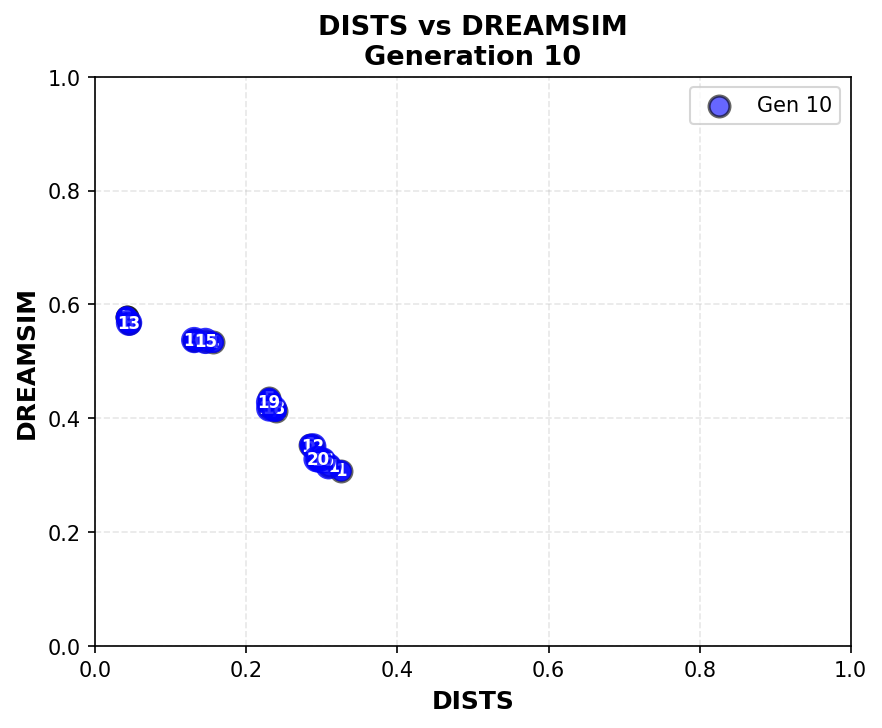}
    \includegraphics[width=0.3\linewidth]{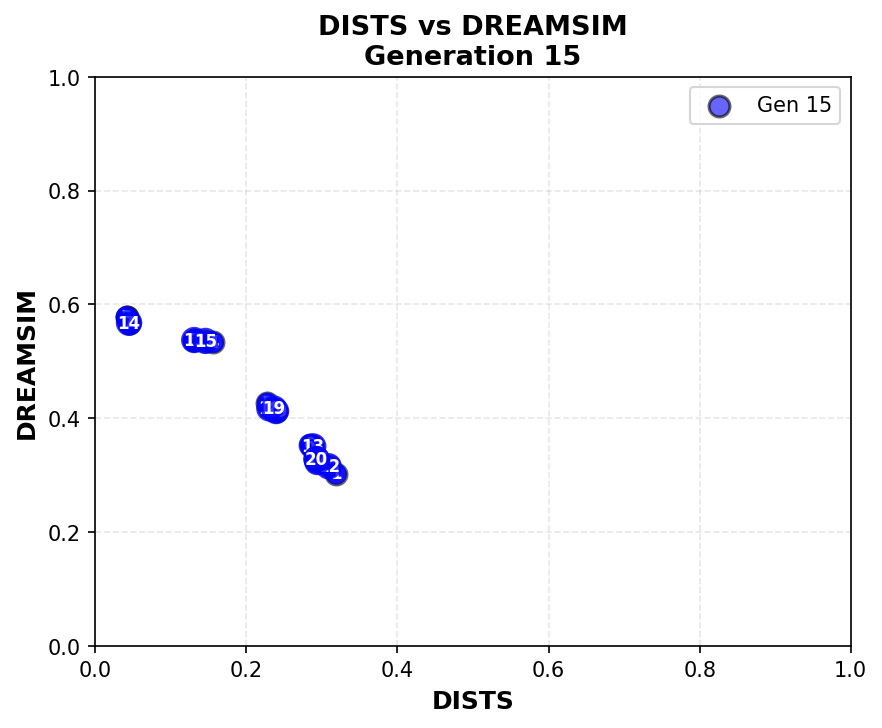}
    \includegraphics[width=0.3\linewidth]{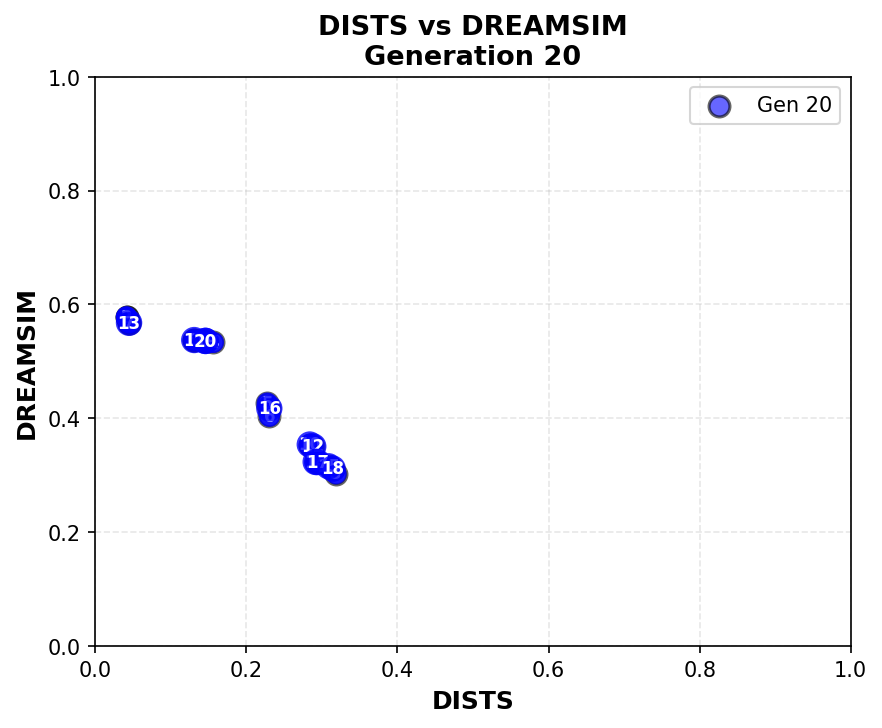}
    \caption{Populations produced at several stages of the optimization process, from generation 1 on the top left to generation 20 on the bottom right. 
    On the $x$ axis is the DISTS value representing content preservation, and on the $y$ axis the DreamSim value representing style transfer. 
    Both metrics range from 1.0 to 0.0, with lower values being better.}
    \label{fig:generations}
    \vspace{-0.3cm}
\end{figure}

\Cref{fig:generations} presents the progression of optimization across twenty generations.
The evolutionary algorithm successfully drives the population toward the Pareto front, with individuals moving from high DISTS and DreamSim values toward the region where both objectives are minimized.

The initial population shows diversity, with solutions distributed across a wide range of combinations.
Several solutions from the first generation are already close to what will become the final Pareto front, which is not surprising due to the relatively small search space.
By generation 3, the characteristic shape of the Pareto front becomes visible.

From generation 5 onward, the population exhibits limited further evolution.
Solutions remain clustered in approximately four to five distinct regions along the Pareto front.
This stagnation after relatively few generations confirms that the search space is small enough to be thoroughly explored with a modest computational budget.

The observed stagnation demonstrates that the problem formulation and resolution method are capable of quickly identifying high-quality solutions, making the approach computationally feasible.
The final Pareto front provides solutions spanning different points along the content-style trade-off curve, enabling practitioners to select pipelines according to specific requirements.

\section{Analysis of Optimized Pipelines}

\subsection{Optimized Pipelines}

\begin{figure}
\vspace{-0.3cm}
\centering
\includegraphics[width=0.45\linewidth]{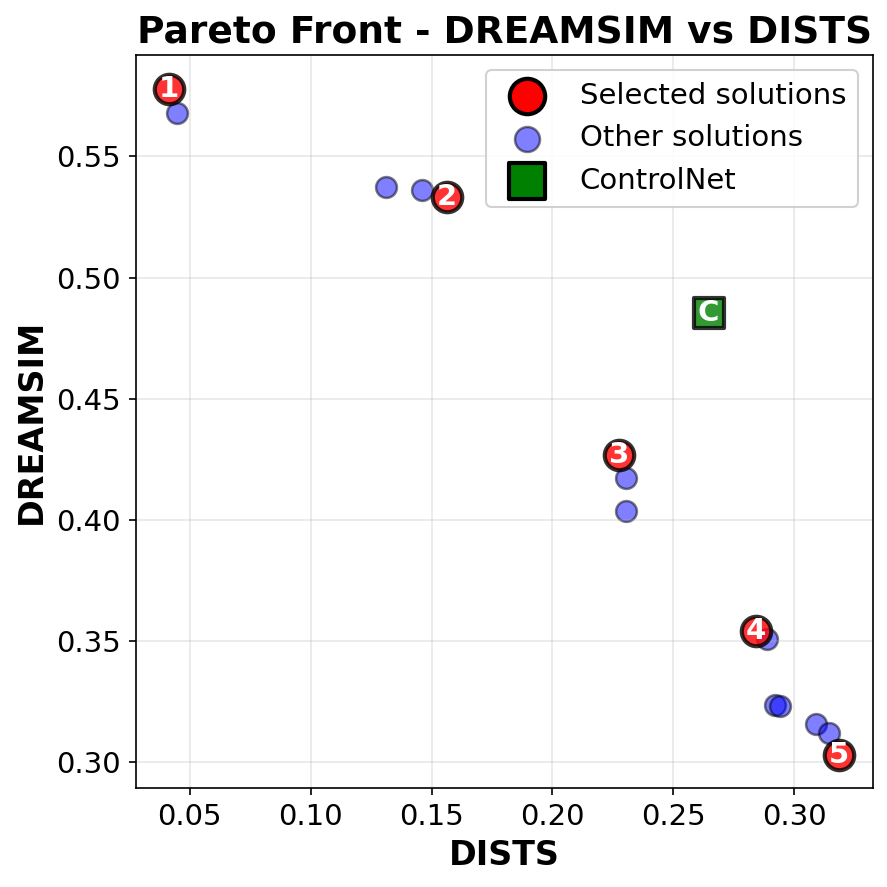}
\caption{The final Pareto front of sequences optimized by NSGA-II. 
The five selected sequences (in red) are highlighted among all optimal solutions (in blue). 
A reference sequence using only ControlNet is shown in green. 
The top-left solution corresponds to the highest content preservation (low DISTS), whereas the bottom-right solution achieves the most realistic style (low DreamSim).}
\label{fig:pareto}
\vspace{-0.3cm}
\end{figure}

Optimization produces a Pareto front containing multiple non-dominated solutions.
To evaluate these solutions, we select representative pipelines that span the front.
We manually select one pipeline from each cluster, resulting in five representative solutions.
\Cref{fig:pareto} shows these selected pipelines in red among the full set of optimal solutions.
The selection ensures even distribution across the trade-off between content preservation and style transfer. For comparison, we also evaluate a baseline pipeline using only ControlNet:
\begin{enumerate}
    \item[C.] ControlNet $\rightarrow$ Stop
\end{enumerate}

The five selected pipelines, ordered from high content preservation to high style fidelity, are:
\begin{enumerate}
    \item Sharpen $\rightarrow$ Stop
    \item Darken $\rightarrow$ Sharpen $\rightarrow$ AdaIN $\rightarrow$ Stop
    \item AdaIN $\rightarrow$ Darken $\rightarrow$ CACTI $\rightarrow$ Stop
    \item ControlNet $\rightarrow$ CACTI $\rightarrow$ Darken $\rightarrow$ Sharpen $\rightarrow$ AdaIN $\rightarrow$ Stop
    \item AdaIN $\rightarrow$ Normalize $\rightarrow$ Blur $\rightarrow$ Contrast $\rightarrow$ ControlNet $\rightarrow$ CACTI $\rightarrow$ Stop
\end{enumerate}

These pipelines reveal patterns about the relationship between solution complexity and objective values.
As DreamSim values decrease (indicating better style transfer), pipelines become progressively more complex.
Pipeline 1 applies only a single operation, while Pipeline 5 uses six of the nine available augmentations.
This suggests that achieving strong style transfer while maintaining reasonable content preservation requires composing multiple transformations.

\begin{figure}[!h]
    \vspace{-0.3cm}
    \centering
    \setlength{\tabcolsep}{2pt} 
    \renewcommand{\arraystretch}{0.9} 
    \begin{tabular}{c c c}
        & Night & Snow \\
        \rotatebox{90}{\parbox[c]{1.7cm}{\centering Reference}} &
        \includegraphics[width=0.29\linewidth]{solutions_fig/content_night.png} &
        \includegraphics[width=0.29\linewidth]{solutions_fig/content_snow.png} \\
        \rotatebox{90}{\parbox[c]{1.7cm}{\centering 1}} &
        \includegraphics[width=0.29\linewidth]{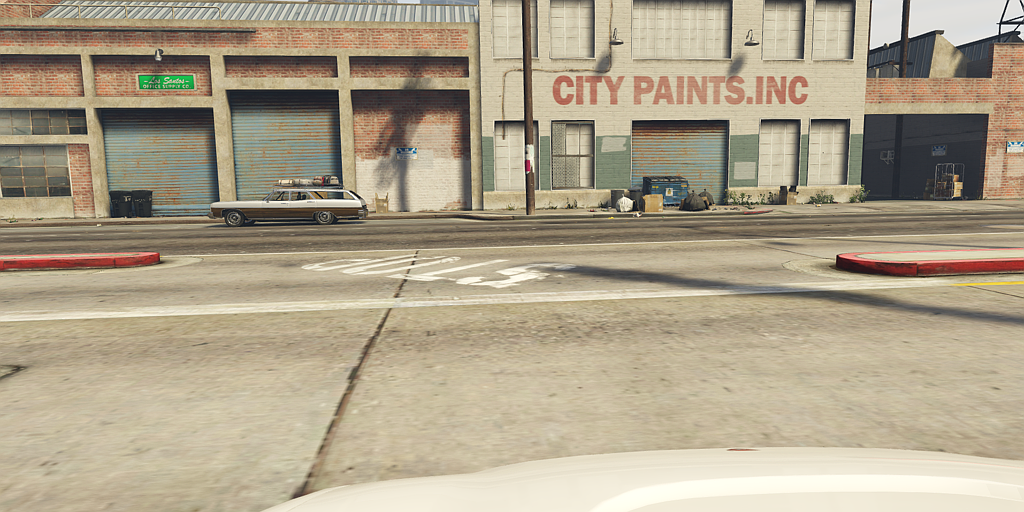} &
        \includegraphics[width=0.29\linewidth]{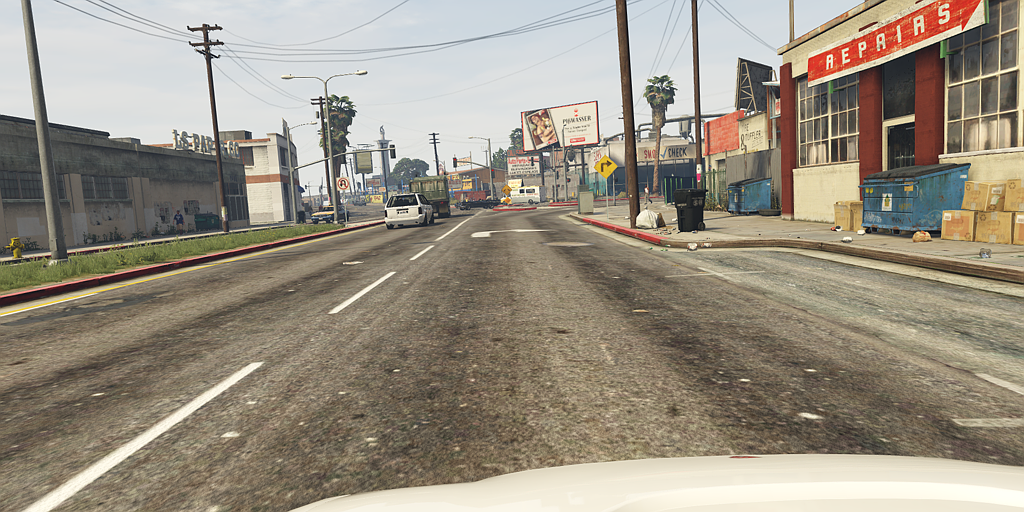} \\
        \rotatebox{90}{\parbox[c]{1.7cm}{\centering 2}} &
        \includegraphics[width=0.29\linewidth]{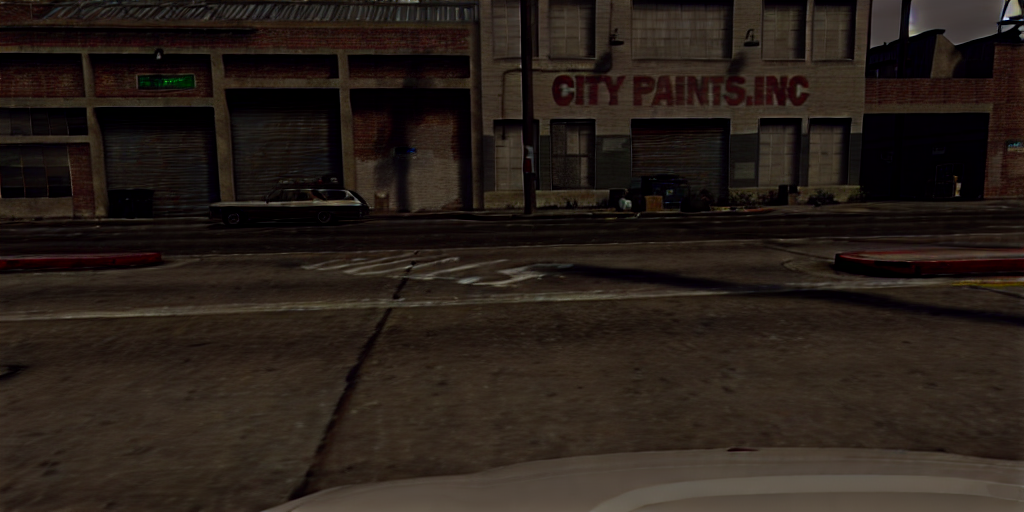} &
        \includegraphics[width=0.29\linewidth]{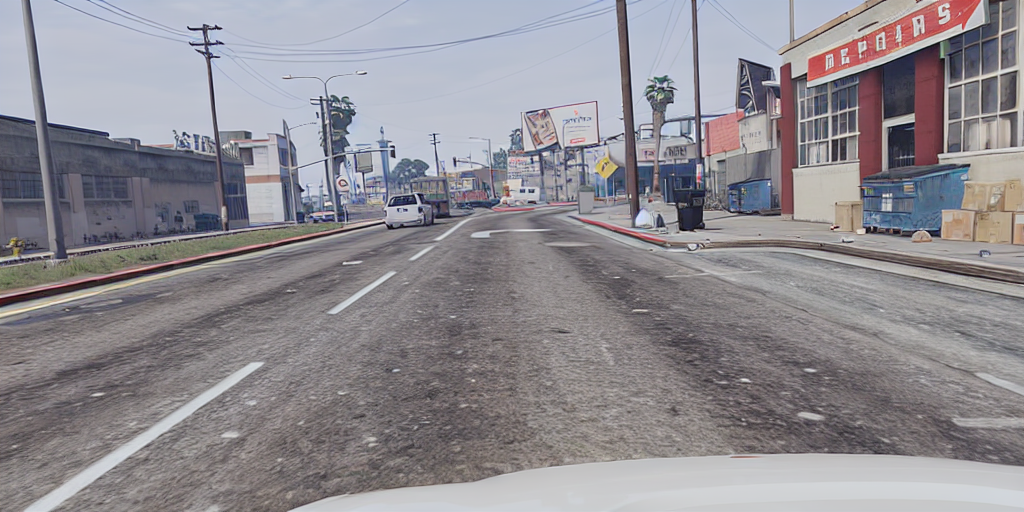} \\
        \rotatebox{90}{\parbox[c]{1.7cm}{\centering 3}} &
        \includegraphics[width=0.29\linewidth]{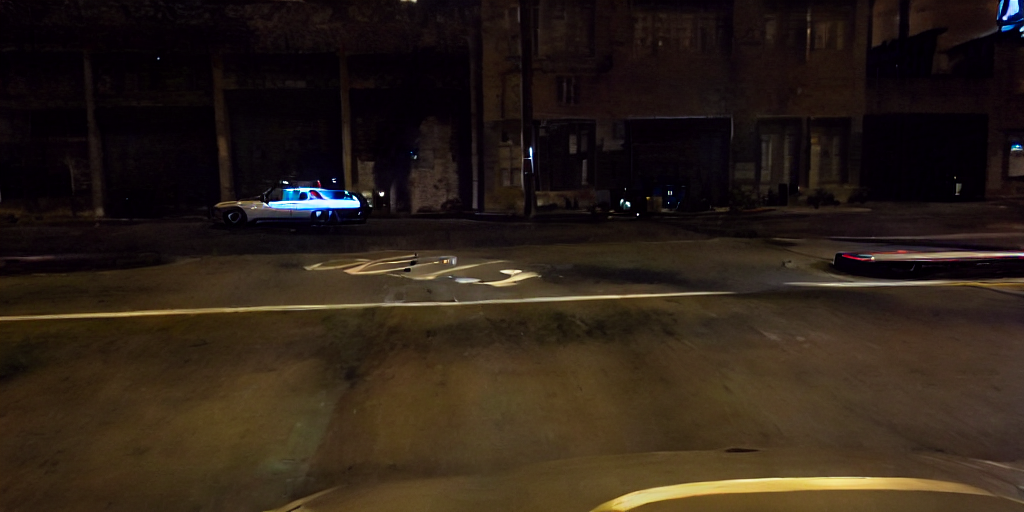} &
        \includegraphics[width=0.29\linewidth]{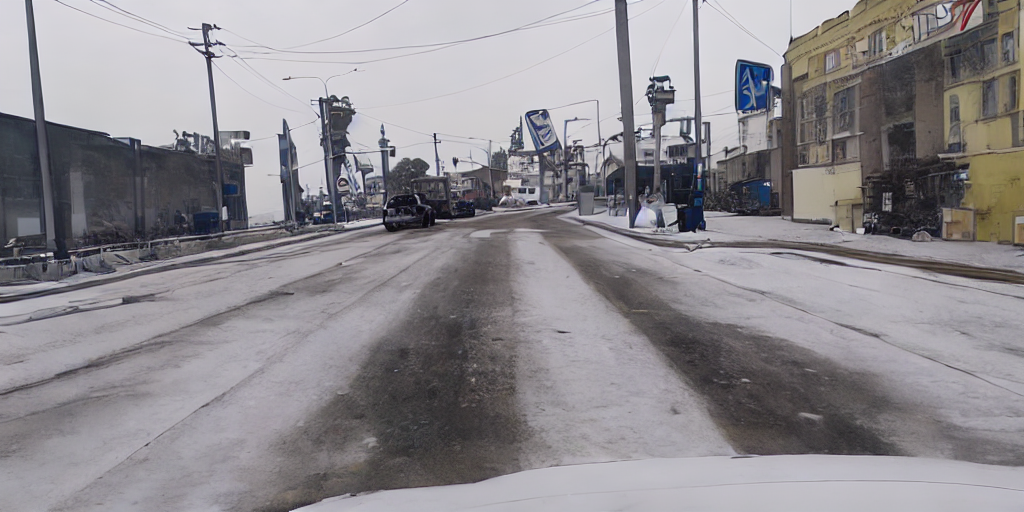} \\
        \rotatebox{90}{\parbox[c]{1.7cm}{\centering 4}} &
        \includegraphics[width=0.29\linewidth]{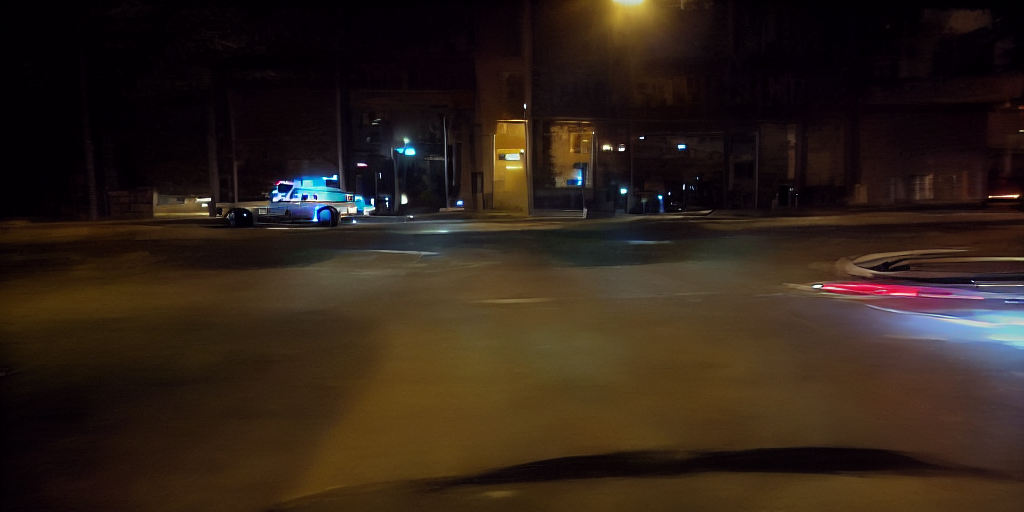} &
        \includegraphics[width=0.29\linewidth]{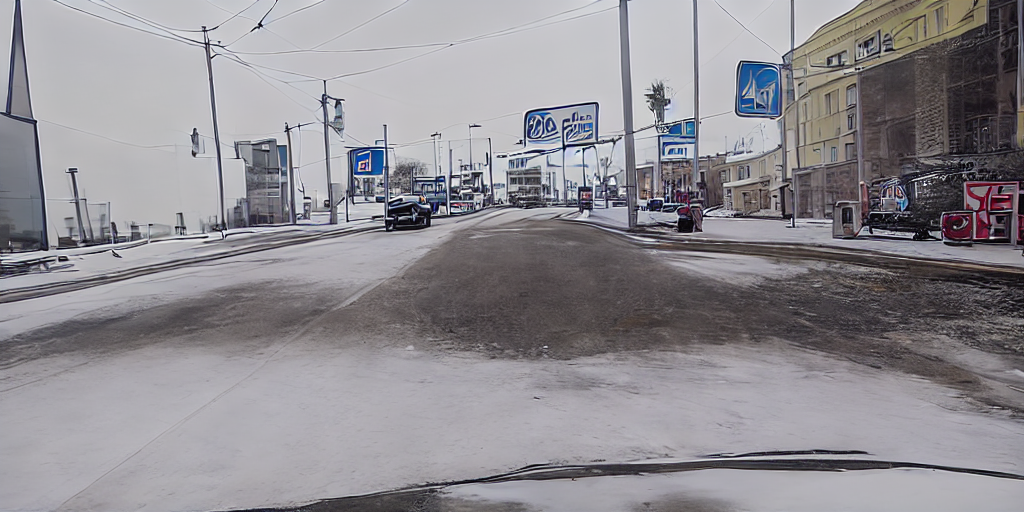} \\
        \rotatebox{90}{\parbox[c]{1.7cm}{\centering 5}} &
        \includegraphics[width=0.29\linewidth]{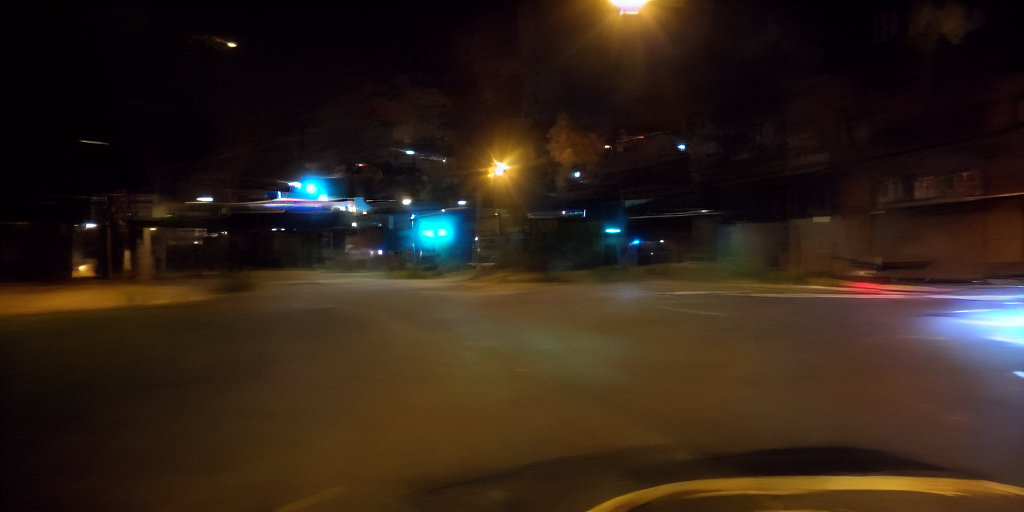} &
        \includegraphics[width=0.29\linewidth]{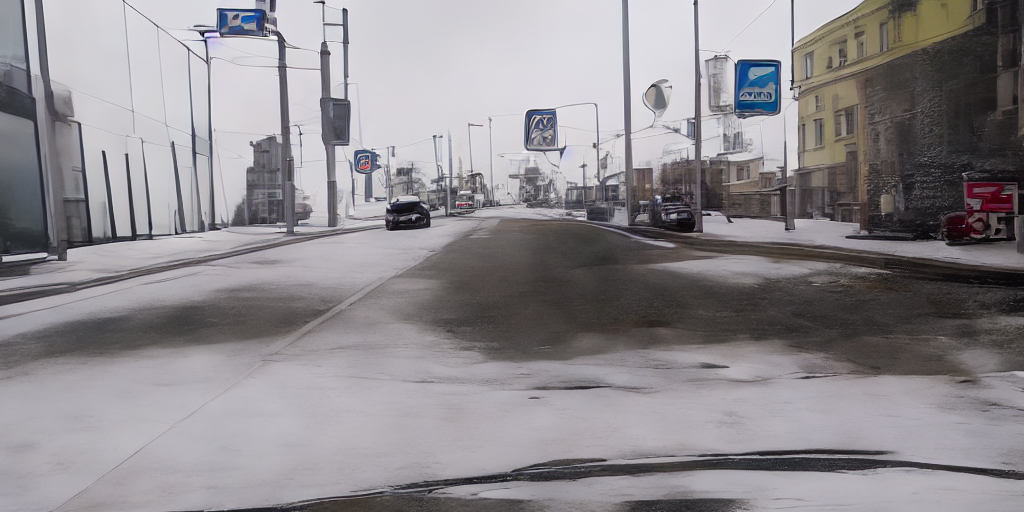} \\
        \rotatebox{90}{\parbox[c]{1.7cm}{\centering ControlNet}} &
        \includegraphics[width=0.29\linewidth]{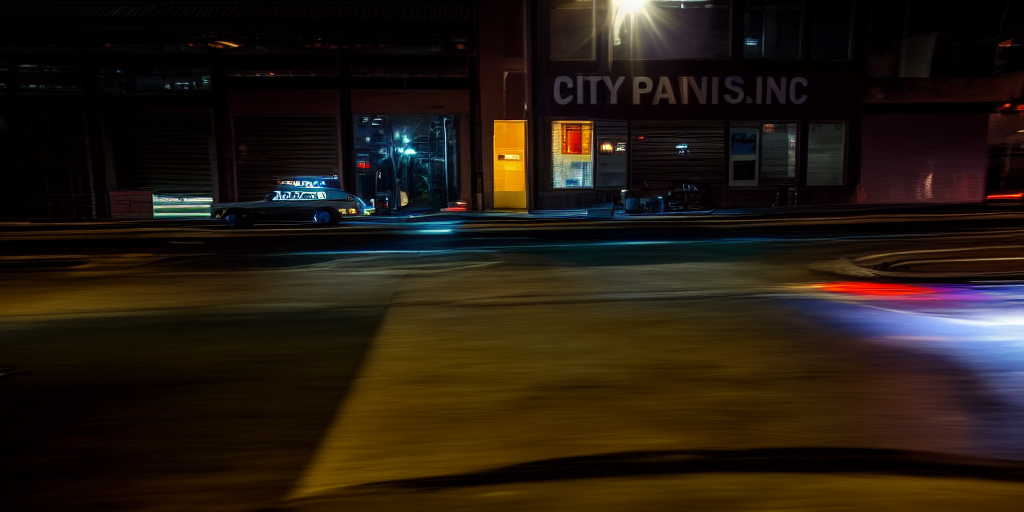} &
        \includegraphics[width=0.29\linewidth]{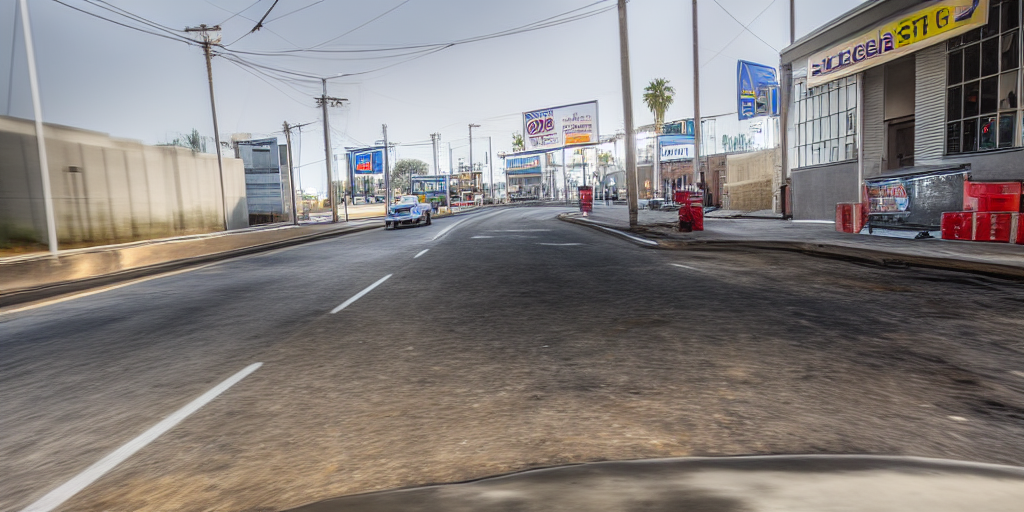} \\
    \end{tabular}
    \caption{Reference content images, then pipelines 1 $\rightarrow$ 5, and finally ControlNet on the Night (left) and Snow (right) reference style images.}
    \label{fig:ex_night}
    \vspace{-0.3cm}
\end{figure}

The presence of deep learning operators AdaIN, ControlNet, and CACTI in the more style-focused pipelines reflects their effectiveness for this task.
ControlNet appears in only two of the five selected pipelines, both of which have DreamSim values in the lower range.
This distribution suggests that ControlNet contributes more to the style transfer objective in this context, which is surprising given its intended effect of content preservation~\cite{zhang2023adding}.


The visual results from the five pipelines presented in~\Cref{fig:ex_night} confirm the progression suggested by DreamSim metric values.
Visual results for the other style images can be found in \Cref{fig:supplementary}.
Across all three weather conditions, images become progressively more stylized from Pipeline 1 to Pipeline 5, with visual characteristics increasingly matching the style reference images.

Pipeline 1, which applies only the sharpen operator, produces images very similar to original GTA5 content.
The primary visible change is enhanced edge definition, with little modification to color distribution, lighting, or texture.
This minimal transformation reveals a limitation of relying solely on DISTS.
DISTS is designed to measure perceptual similarity with emphasis on structural content, making it relatively insensitive to subtle stylistic differences between synthetic and real domains.

Images from Pipelines 2 through 5 show increasingly substantial modifications to visual appearance while maintaining recognizable scene structure.
Colors shift to match real-world distributions, lighting conditions become less uniform, and texture characteristics change from smooth rendered appearance to more complex patterns of photographic imagery.
These changes are what DreamSim is designed to capture: perceptual style similarity that encompasses broader visual characteristics distinguishing synthetic and real images.

The balance between DISTS and DreamSim becomes apparent through these visual results.
DISTS prevents pipelines from applying transformations so aggressive that spatial structure is lost, which would invalidate ground truth annotations.
DreamSim encourages sufficient style modification to reduce the domain gap.
Pipelines in the middle of the Pareto front (particularly Pipelines 3 and 4) achieve a reasonable compromise.

On the other hand, the ControlNet outputs differ noticeably from those produced by the selected pipelines. Although the overall structure looks well preserved, the generated images tend to have an overly smooth texture that drifts further from the style reference. In the night scenario, ControlNet achieves a reasonably strong style transfer, producing a dark scene with pronounced lighting and a fading of the road markings. In contrast, the snow scenario shows only minimal stylistic modification, with slight adjustments in lighting and color. Compared to Pipeline 3, the ControlNet results appear to have higher perceptual quality, yet they exhibit weaker alignment with the style reference and less faithful preservation of content details such as building structure.

\subsection{Pipeline Evaluation}

\setlength{\tabcolsep}{8pt}
\begin{table}
    \vspace{-0.3cm}
\centering
\begin{tabular}{l c c c c c c c}
 & DISTS & DISTS & DreamSim & DreamSim & CMMD \\
Samples & 5 & 1250 & 5 & 1250 & 1250 \\
\toprule
Pipeline 1 & 0.0415 & 0.0396 & 0.5779 & 0.5600 & 5.15\\ 
\hline
Pipeline 2 & 0.1563 & 0.1478 & 0.5334 & 0.5241 & 4.57 \\ %
\hline
Pipeline 3 & 0.2277 & 0.2249 & 0.4266 & 0.3837 & 2.67 \\ 
\hline
Pipeline 4 & 0.2840 & 0.2777 & 0.3540 & 0.3465 & 2.37 \\ 
\hline
Pipeline 5 & 0.3185 & 0.3025 & \textbf{0.3030} &\textbf{0.3177} & \textbf{2.35} \\ 
\midrule[1pt]
GTA5 & \textbf{0.0000} & \textbf{0.0000} & 0.5766 & 0.5577  & 5.18 \\
\hline
ControlNet & 0.2646 & 0.2666 & 0.4852 & 0.4653 & 3.48 \\ 
\bottomrule\\
\end{tabular}
\caption{Image quality metrics for the five selected pipelines and the baseline, computed between content and style pairs of reference images, as well as generated datasets (1250 images) and the target domain. 
}
\label{tab:quality_results}
    \vspace{-0.3cm}
\end{table}

For each selected pipeline, we generate a complete dataset by applying the augmentation sequence to source domain images.
We generate 250 transformed images for each of the five target weather conditions, producing a total dataset of 1250 images.


\Cref{tab:quality_results} reports the DISTS and DreamSim values used as optimization metrics in \Cref{fig:pareto}, computed both on the small evaluation set (5 images) and on the full generated datasets (1250 images per pipeline).
Across all pipelines, DISTS and DreamSim remain highly consistent when moving from 5 to 1250 images. This stability indicates that both metrics are robust to dataset scaling and can therefore be reliably estimated from only a few samples during optimization.
As DreamSim improves and DISTS gets worse (moving from Pipeline 1 to Pipeline 5), CMMD generally decreases, indicating better distributional match.
This correlation between DreamSim and CMMD further validates the use of pairwise metrics in the proposed optimization approach.

\setlength{\tabcolsep}{8pt}
\begin{table}
    \vspace{-0.3cm}
\centering
\begin{tabular}{ l c c c c c c c }
 & Fog & Night & Rain & Snow & ACDC & Cityscapes \\
\toprule
Pipeline 1 & 48.57 & 16.65 & 42.50 & 40.34 & 37.07 & 51.41 \\
\hline
Pipeline 2 & 50.65 & 19.89 & 45.65 & 43.54 & 39.40 & 51.51  \\
\hline
Pipeline 3 & 49.99 & 19.32 & 45.17 & 42.55 & 39.79 & 50.69 \\
\hline
Pipeline 4 & 50.94 & 19.46 & 45.94 & 44.02 & 40.27 & 51.36 \\
\hline
Pipeline 5 & 48.91 & 17.57 & 44.29 & 42.69 & 38.73 & 50.09 \\
\midrule[1pt]
GTA5 & 48.74 & 16.35 & 42.52 & 40.33 & 36.64 & 50.56 \\
\hline
ControlNet & \textbf{55.73} & \textbf{24.84} & \textbf{46.66} & \textbf{44.39} & \textbf{42.42} & \textbf{52.12} \\
\bottomrule\\
\end{tabular}
\caption{Evaluation metrics for the five selected pipelines and the baseline. 
mIoU values are shown for each ACDC weather condition, the mean across ACDC conditions and Cityscapes clear day images.}
\label{tab:eval_results}
    \vspace{-0.3cm}
\end{table}

Next, we evaluate the generated data on semantic segmentation model training to evaluate the capacity of these style transfer pipelines to aid in domain adaptation.
We train DAFormer~\cite{hoyer2022daformer}, a transformer-based domain adaptation architecture, on each generated dataset.
Following standard protocols~\cite{hoyer2022daformer,benigmim2023datum,jia2024dginstyle,chigot2025cactif}, we train for 40,000 iterations with batch size 2.
After training, we evaluate each model on ACDC and Cityscapes validation sets.
We report mean Intersection over Union (mIoU), the pixel-wise segmentation accuracy averaged across all semantic classes.
This metric directly reflects downstream task performance achieved by training on datasets generated with each optimized pipeline.

\Cref{tab:eval_results} presents downstream task performance of each pipeline.
mIoU scores vary across pipelines and weather conditions.
Furthermore, the baseline ControlNet achieves mIoU values that exceed most optimized pipelines across several conditions.
Having been fine-tuned directly on the style images, ControlNet can effectively combine the learned target-domain features with strong class and structural conditioning. Even though its outputs show weaker perceptual similarity, particularly in textures and lighting, ControlNet generates images that stay close to the target domain in terms of semantic content. This strong semantic coherence helps explain its comparatively high downstream performance despite less accurate style transfer.

The segmentation results indicate that distributional similarity between generated and target data does not directly correlate with training performance.
This finding aligns with recent work on domain generalization, particularly DATUM~\cite{benigmim2023datum}, which demonstrates that realistic appearance is not necessary for training segmentation networks.

\section{Conclusion}

This work demonstrates that evolutionary algorithms can effectively explore the space of data augmentation pipelines for domain adaptation.
The proposed TSP formulation allows multi-objective optimization using pairwise metrics to generate multiple candidate solutions.
This formulation constitutes a novel application of evolutionary computation to automate the design of style transfer pipelines for semantic segmentation domain adaptation.
As measured by DISTS and DreamSim, our approach produces pipelines with good style transfer properties and diverse trade-offs between content preservation and style matching.

A key advantage of this method is computational efficiency.
Optimizing the complete set of pipelines requires the same time as evaluating a single pipeline for full data generation.
This means the quick metrics can be applied to a new target domain before committing to expensive data generation, allowing practitioners to consider multiple pipeline options early in the adaptation process.

However, our results reveal a limitation in the choice of objectives.
While the evolved pipelines achieve good style transfer, they do not match the downstream segmentation performance of ControlNet alone when measured by mIoU.
This indicates that the DISTS and DreamSim combination, while effective for assessing style transfer quality, does not serve as reliable surrogates for downstream task performance on image segmentation.
Future work should focus on identifying metrics that can cheaply estimate downstream segmentation quality.
For domain adaptation specifically, an ideal formulation would include three objectives: DreamSim for style matching, a surrogate metric for segmentation performance, and compute time.
Additional directions include expanding the set of available augmentation operations and incorporating hyperparameter tuning directly into the genetic algorithm.



%
%
%
\bibliographystyle{splncs04}
\bibliography{biblio}
\newpage

\section{Appendix}
\begin{figure}
    \vspace{-0.3cm}
    \centering
    \setlength{\tabcolsep}{2pt}
    \renewcommand{\arraystretch}{0.9}
    \begin{tabular}{c c c c}
        & Fog & Rain & Day \\
        \rotatebox{90}{\parbox[c]{1.7cm}{\centering Reference}} &
        \includegraphics[width=0.29\linewidth]{solutions_fig/content_fog.png} &
        \includegraphics[width=0.29\linewidth]{solutions_fig/content_rain.png} &
        \includegraphics[width=0.29\linewidth]{solutions_fig/content_day.png} \\

        \rotatebox{90}{\parbox[c]{1.7cm}{\centering 1}} &
        \includegraphics[width=0.29\linewidth]{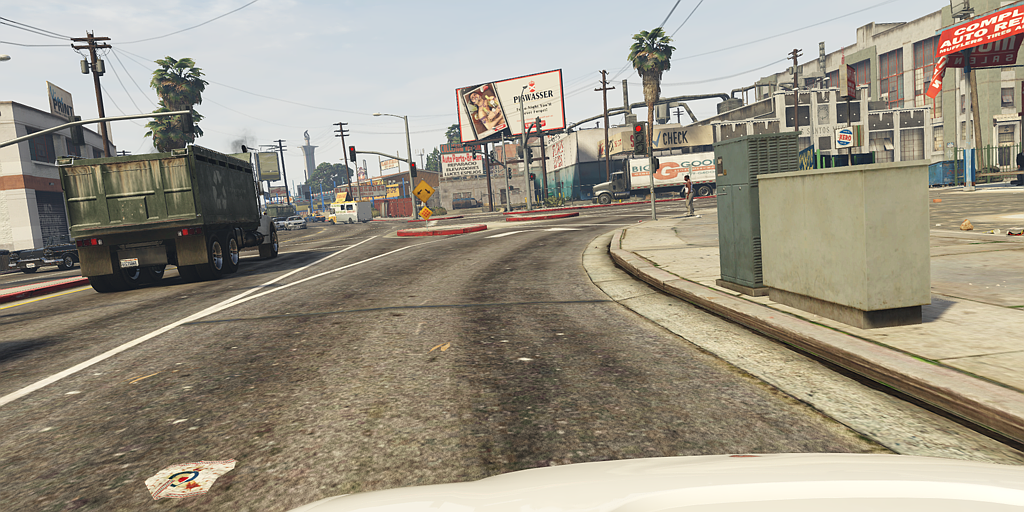} &
        \includegraphics[width=0.29\linewidth]{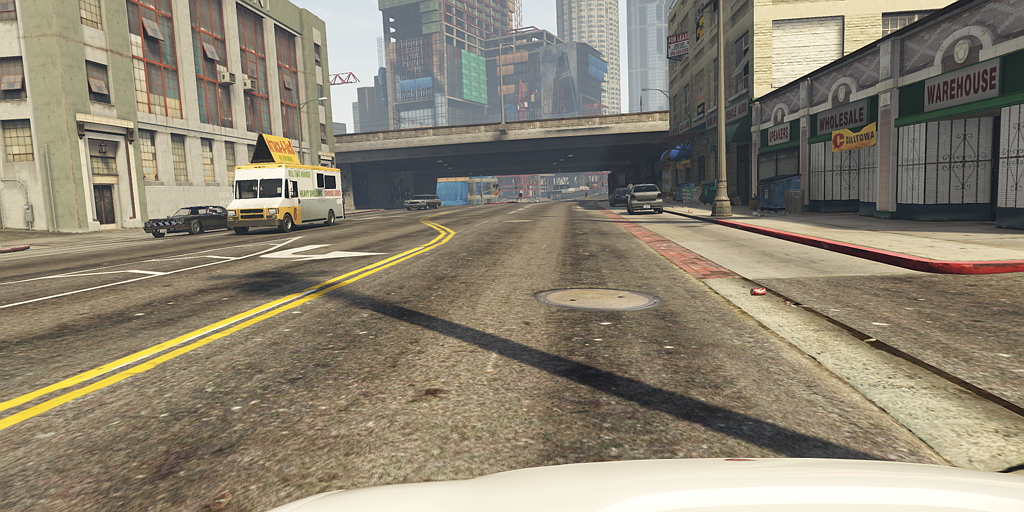} &
        \includegraphics[width=0.29\linewidth]{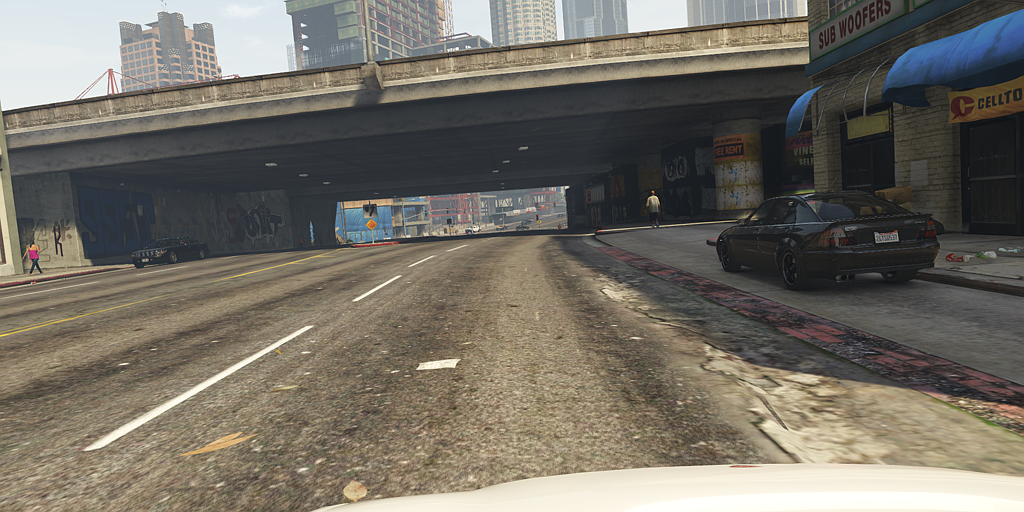} \\

        \rotatebox{90}{\parbox[c]{1.7cm}{\centering 2}} &
        \includegraphics[width=0.29\linewidth]{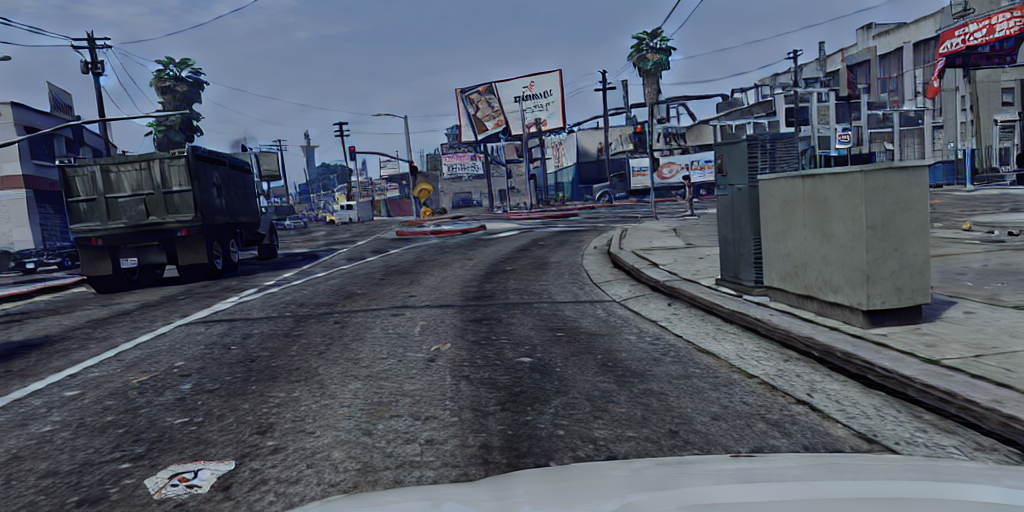} &
        \includegraphics[width=0.29\linewidth]{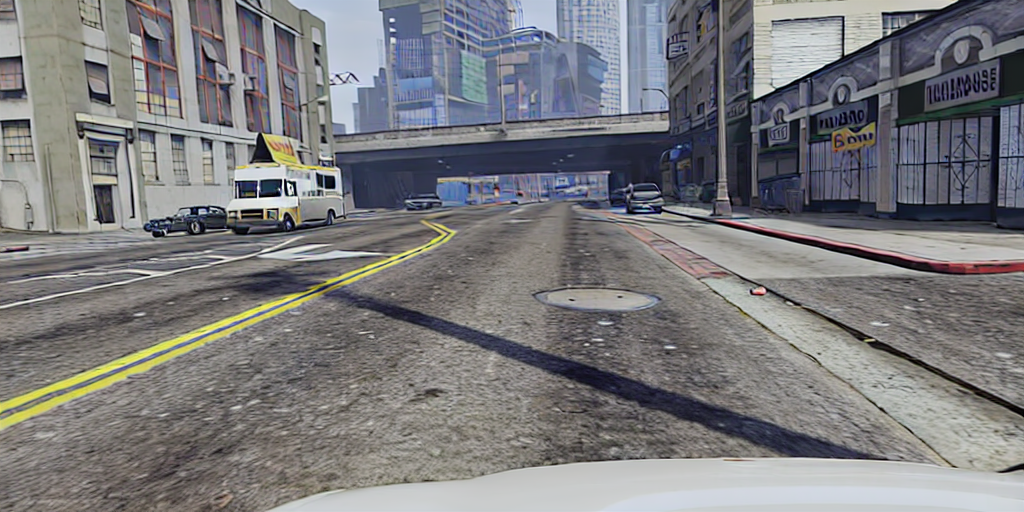} &
        \includegraphics[width=0.29\linewidth]{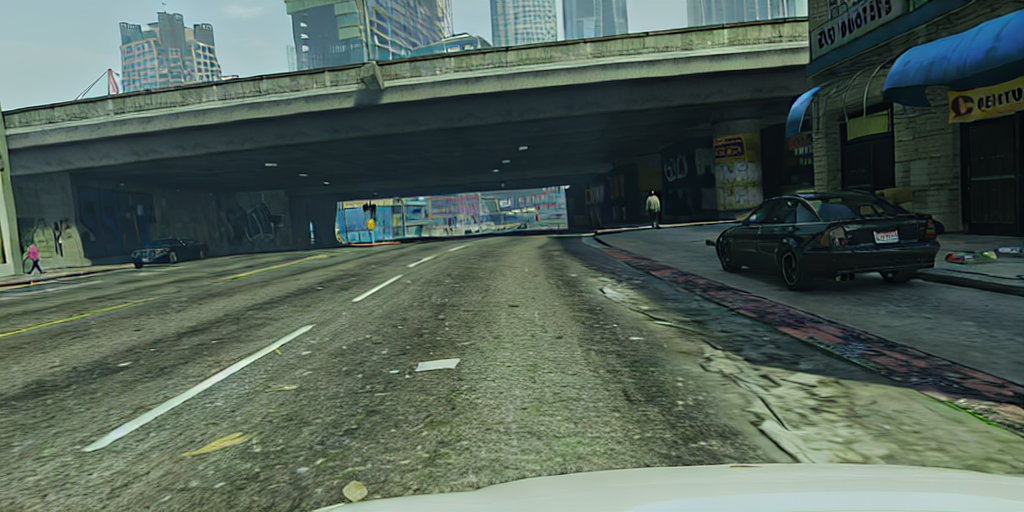} \\

        \rotatebox{90}{\parbox[c]{1.7cm}{\centering 3}} &
        \includegraphics[width=0.29\linewidth]{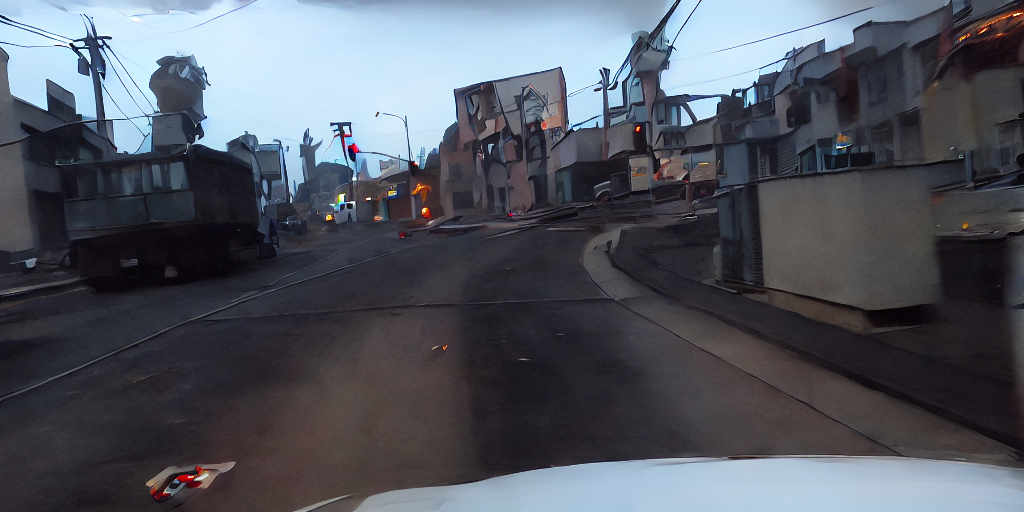} &
        \includegraphics[width=0.29\linewidth]{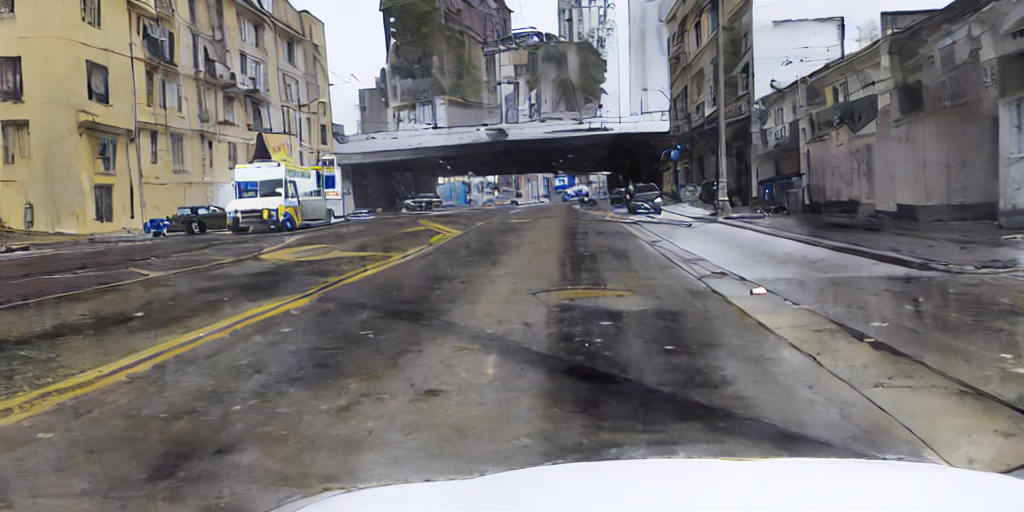} &
        \includegraphics[width=0.29\linewidth]{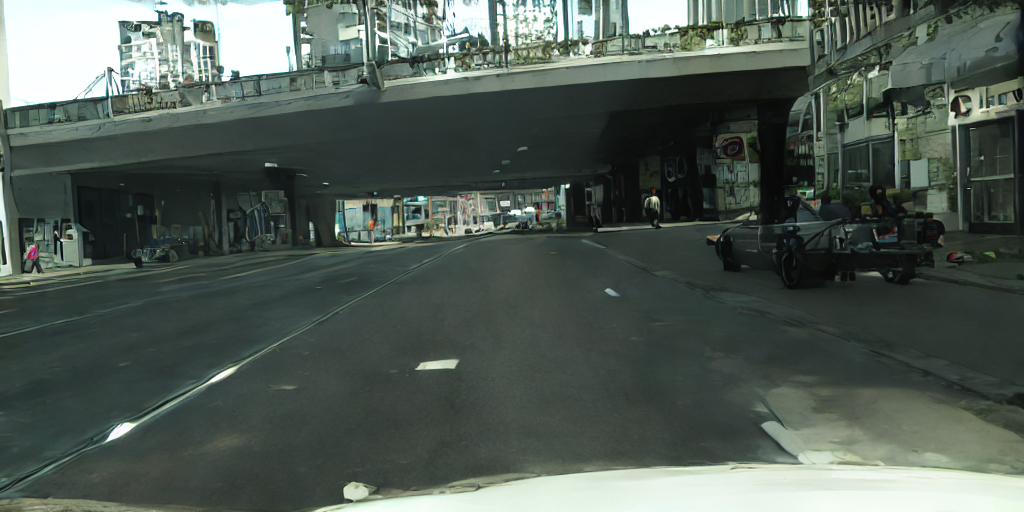} \\

        \rotatebox{90}{\parbox[c]{1.7cm}{\centering 4}} &
        \includegraphics[width=0.29\linewidth]{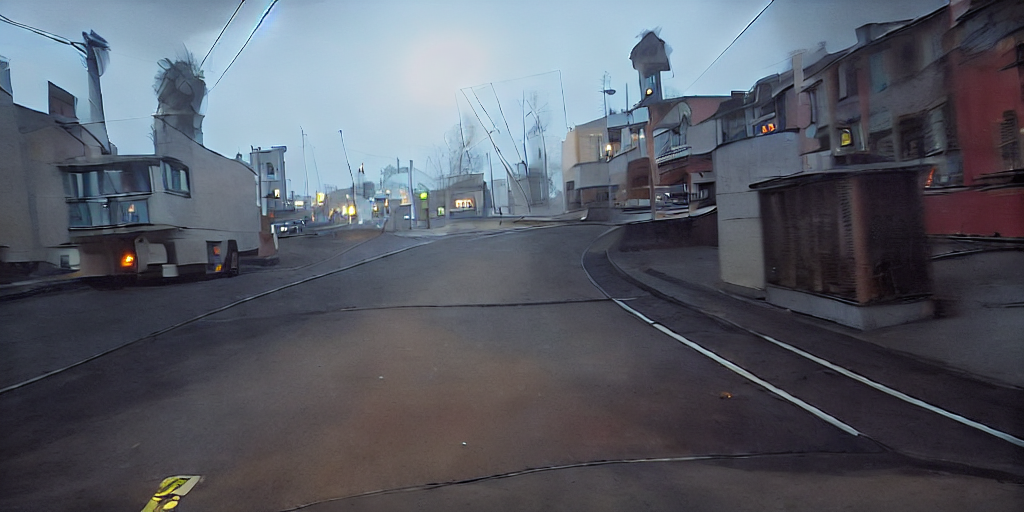} &
        \includegraphics[width=0.29\linewidth]{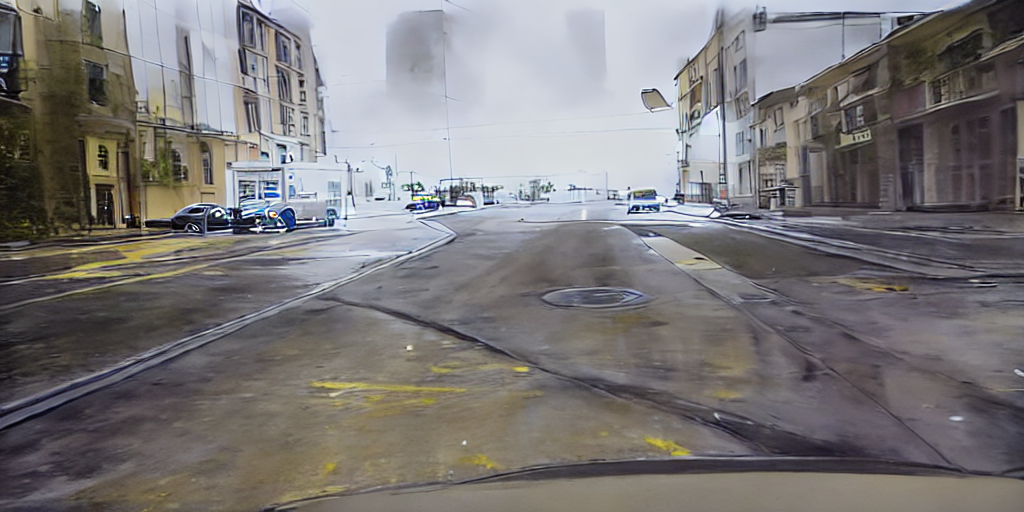} &
        \includegraphics[width=0.29\linewidth]{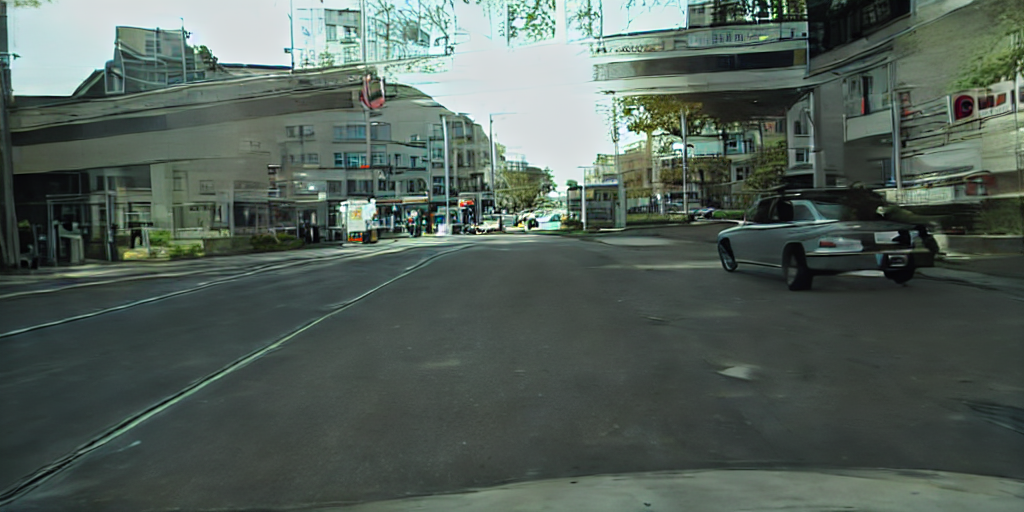} \\

        \rotatebox{90}{\parbox[c]{1.7cm}{\centering 5}} &
        \includegraphics[width=0.29\linewidth]{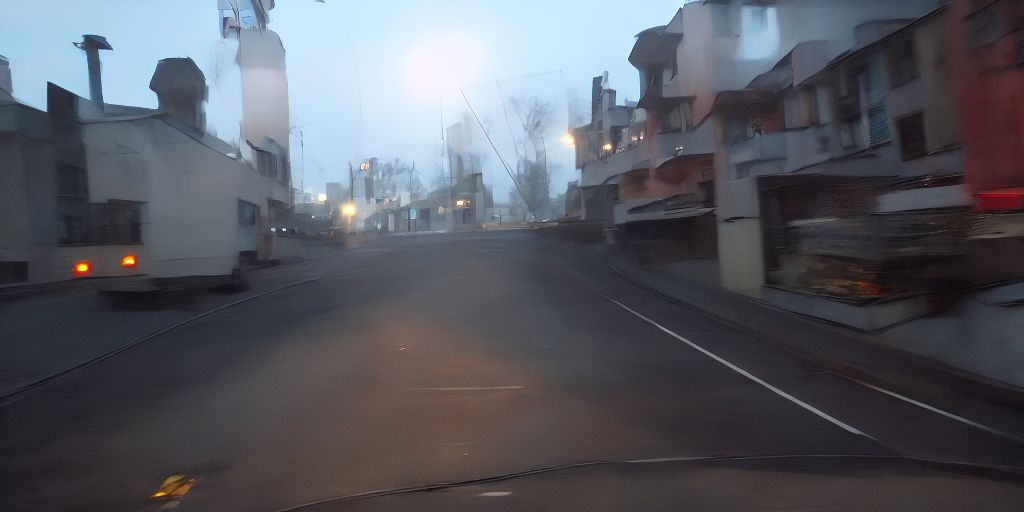} &
        \includegraphics[width=0.29\linewidth]{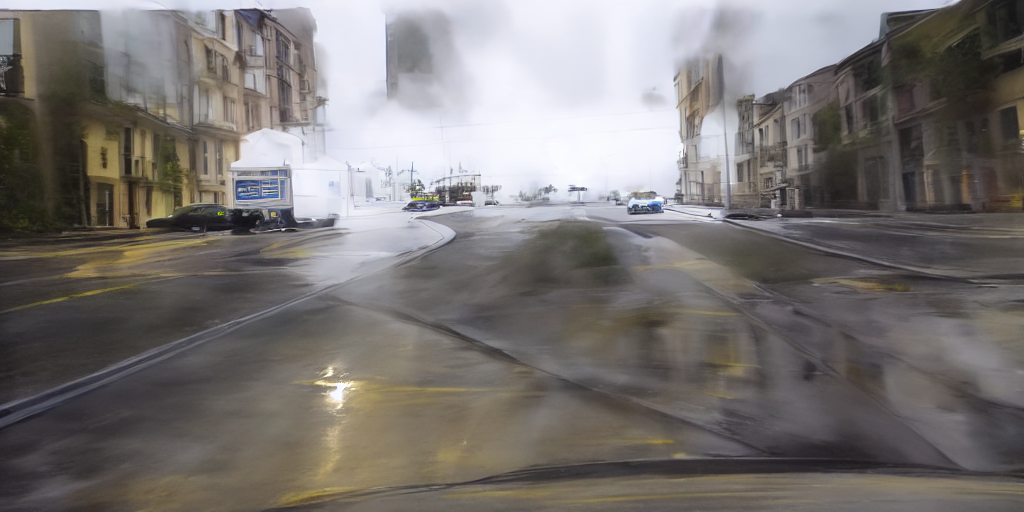} &
        \includegraphics[width=0.29\linewidth]{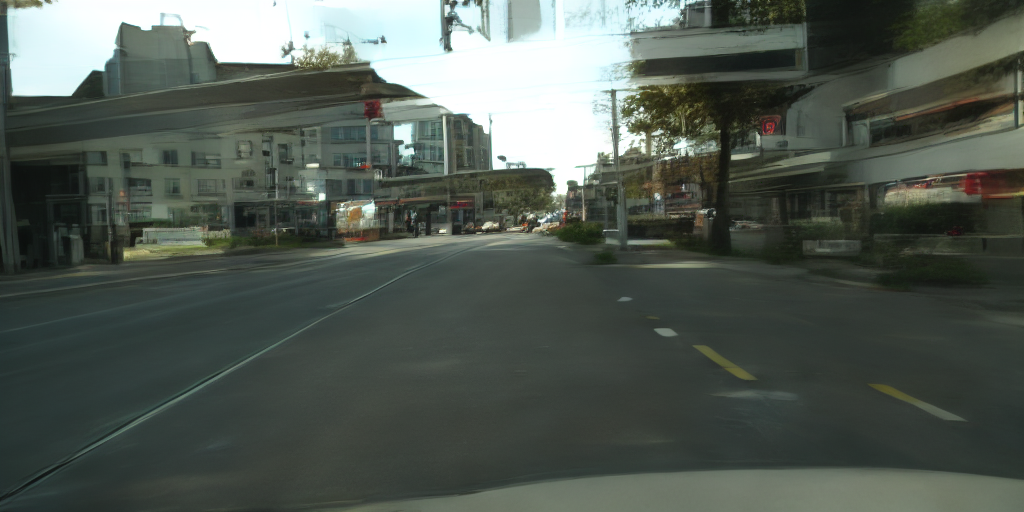} \\

        \rotatebox{90}{\parbox[c]{1.7cm}{\centering ControlNet}} &
        \includegraphics[width=0.29\linewidth]{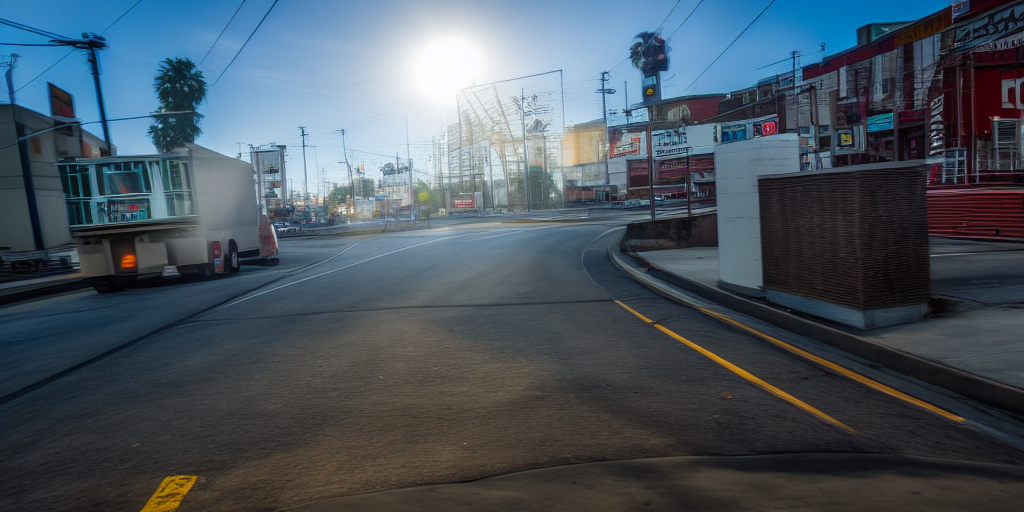} &
        \includegraphics[width=0.29\linewidth]{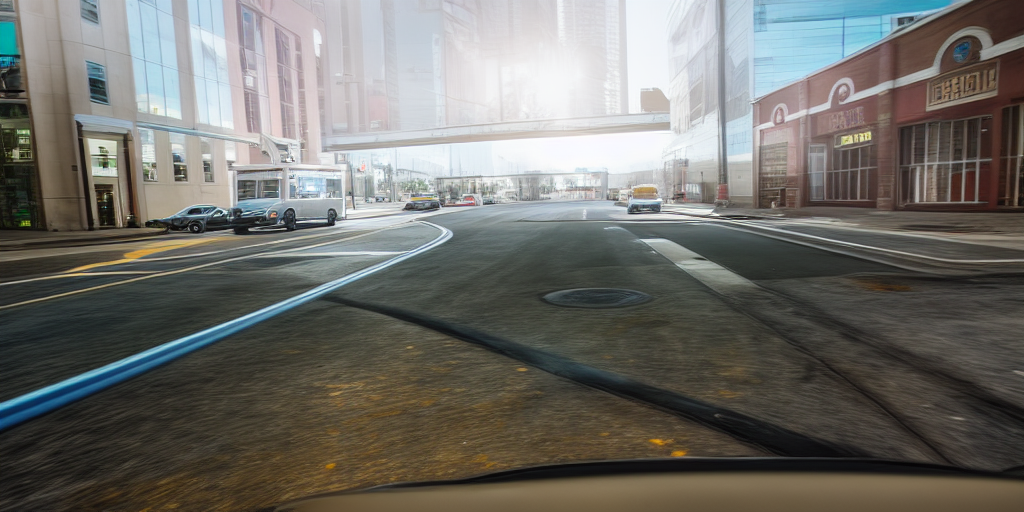} &
        \includegraphics[width=0.29\linewidth]{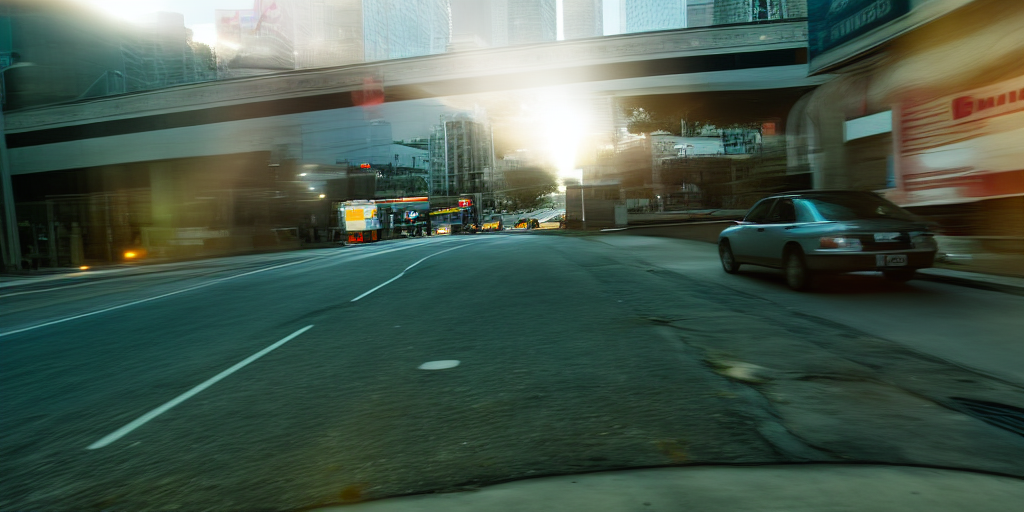} \\
    \end{tabular}
    \caption{Reference content images, then pipelines 1 $\rightarrow$ 5, and finally ControlNet on the Fog, Rain, and Day reference style images.}
    \label{fig:supplementary}
    \vspace{-0.3cm}
\end{figure}

\end{document}